\documentclass[letterpaper]{article} 
\usepackage{aaai24}  
\usepackage{times}  
\usepackage{helvet}  
\usepackage{courier}  
\usepackage[hyphens]{url}  
\usepackage{graphicx} 
\usepackage{natbib}  
\usepackage{caption} 
\usepackage{subcaption}
\usepackage{amsthm}
\usepackage{tikz}
\usepackage{pgfplots}
\usetikzlibrary{calc}
\usepackage{epsfig}
\usepackage{graphicx}
\usepackage{epstopdf}
\usepackage{amsmath}
\usepackage{amssymb}
\usepackage{booktabs}
\usepackage{caption}
\usepackage{float} 
\usepackage{array}
\usetikzlibrary{calc}
\usepackage{tabularx}
\usepackage{color}
\theoremstyle{definition}

\usepackage{adjustbox}
\usepackage{pifont}
\usepackage{multirow}
\usepackage{multicol}
\usepackage{xcolor}
\newcommand{\ie}{\textit{i.e.}}
\newcommand{\eg}{\textit{e.g.}}

\renewcommand{\paragraph}[1]{\noindent\textbf{#1}}

\usepackage{algorithm}
\usepackage{algorithmic}

\makeatletter
\renewcommand{\copyright@on}{}
\makeatother

\usepackage{newfloat}
\usepackage{listings}
\DeclareCaptionStyle{ruled}{labelfont=normalfont,labelsep=colon,strut=off} 
\floatstyle{ruled}
\newfloat{listing}{tb}{lst}{}
\floatname{listing}{Listing}
\title{LCCo: Lending CLIP to Co-Segmentation}
\author{
    Xin Duan$^{1}$,  \quad
    Yan Yang$^{2}$,  \quad
    Liyuan Pan$^{1}$,  \quad
    Xiabi Liu$^{1}$
}
\affiliations{
    \textsuperscript{\rm 1}Beijing Institute of Technology,  \quad \textsuperscript{\rm 2}Australian National University \\
    \{duanxin, liyuan.pan, liuxiabi\}@bit.edu.cn \quad yan.yang@anu.edu.au
}

\usepackage{bibentry}

\begin{document}

\maketitle

\begin{abstract}
This paper studies co-segmenting the common semantic object in a set of images. Existing works either rely on carefully engineered networks to mine the implicit semantic information in visual features or require extra data (\ie, classification labels) for training. In this paper, we leverage the contrastive language-image pre-training framework (CLIP) for the task. With a backbone segmentation network that independently processes each image from the set, we introduce semantics from CLIP into the backbone features, refining them in a coarse-to-fine manner with three key modules:
i) an image set feature correspondence module, encoding global consistent semantic information of the image set;
ii) a CLIP interaction module, using CLIP-mined common semantics of the image set to refine the backbone feature;
iii) a CLIP regularization module, drawing CLIP towards this co-segmentation task, identifying the best CLIP semantic and using it to regularize the backbone feature.
Experiments on four standard co-segmentation benchmark datasets show that the performance of our method outperforms state-of-the-art methods.

\end{abstract}

\section{Introduction}

\begin{figure}
\begin{center}
    \begin{subtable}{0.9\linewidth}
    \setlength{\tabcolsep}{3pt}
    \begin{tabular}{ccc}
      \includegraphics[width=.31\linewidth]{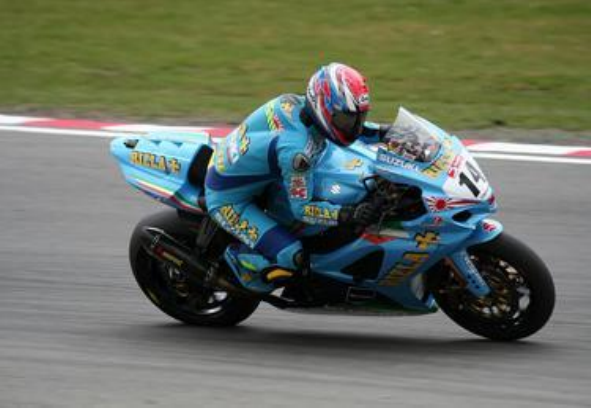}  & \includegraphics[width=.31\linewidth]{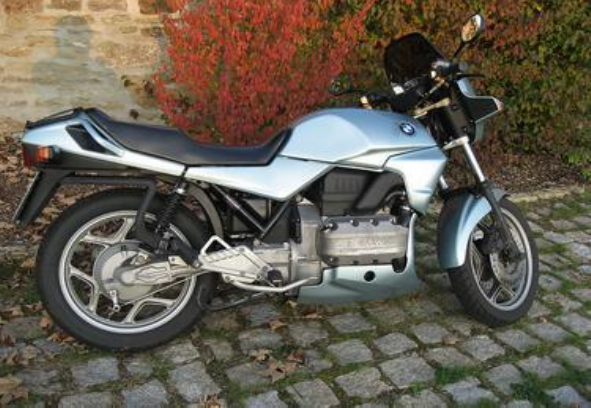}  & \includegraphics[width=.31\linewidth]{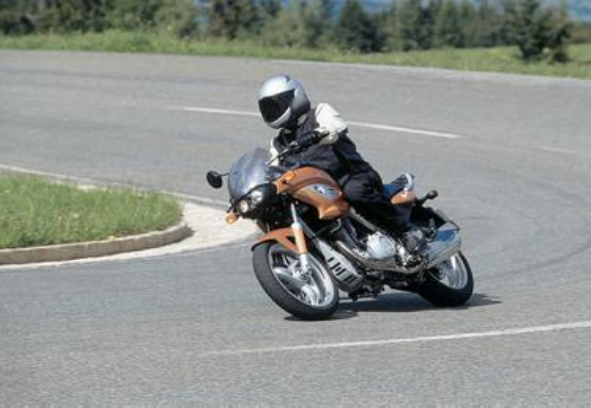}  \\
      \includegraphics[width=.31\linewidth]{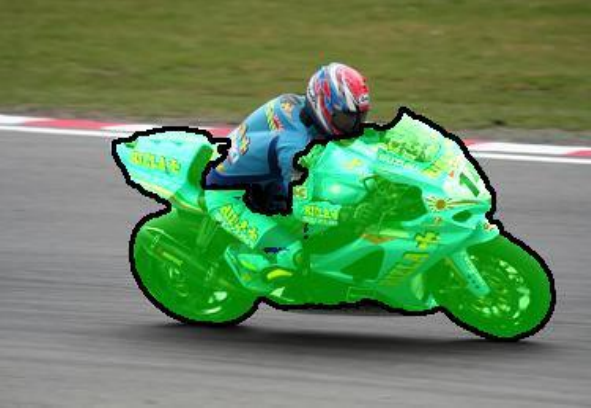}  & \includegraphics[width=.31\linewidth]{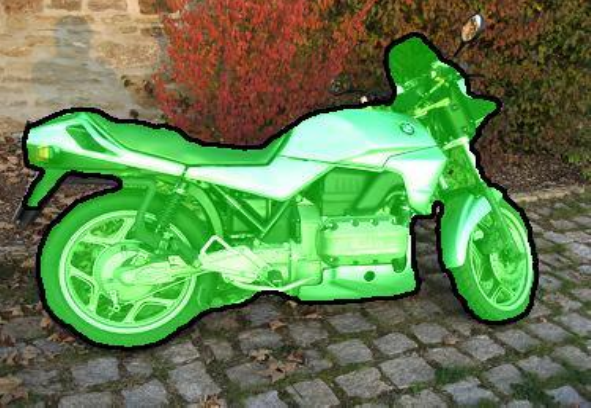}  & \includegraphics[width=.31\linewidth]{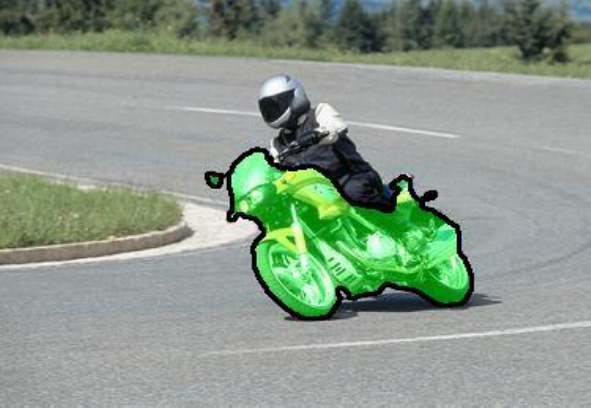}   \\
    \end{tabular}
    \vspace{-.5em}

    \end{subtable} \\

\end{center}
 \caption{\it \small \small Examples of our co-segmentation results. Given an image set (top row), where its common semantic object is `Motorbike', we aim to estimate the semantic mask for each image in the set, corresponding to the common semantic object (bottom row).
 }
\label{horse}
\vspace{-1.5em}
\end{figure}

This paper investigates the problem of image co-segmentation. Given a set of images, we aim to find the common semantic object within the image set and generate segmentation masks for the object in each image. Fig.~\ref{horse} illustrates an example scenario of the co-segmentation problem.
The co-segmentation problem has been well studied for applications of 3D reconstruction \cite{mustafa2017semantically}, image retrieval \cite{shen2022learning}, video salient detection \cite{su2022unified}, image matching \cite{zhang2020deepemd} and video object tracking \cite{liu2020weakly}. 

Previous efforts \cite{banerjee2019cosegnet, chen2018semantic, li2018deep} have been primarily based on Siamese networks to extract image features, and enable feature interaction to identify the common semantics that are implicitly encoded in the visual features for segmenting the object. However, the interaction is not restricted to reasoning the common semantic information, and background noise has also interacted \cite{hsu2018co, sidi2011unsupervised}. Collecting accurate ground-truth common semantic classes \cite{li2018deep, chen2018semantic, zhang2020deep, su2022unified} to supervisedly constrain the feature interaction can mitigate the aliasing phenomenon, yet the problem is not to be addressed and involves extra data in the training phase.

Inspired by the strong semantic discovery ability of the pre-trained contrastive language-image pre-training framework~(CLIP), we propose our \underline{L}ending \underline{C}LIP to \underline{Co}-Segmentation framework, LCCo, that explicitly encodes and exploits common semantic information mined by CLIP for the co-segmentation problem. Besides getting accurate co-segmented masks, powered by the semantic knowledge from the CLIP, we also unlock the accuracy-improving potential with respect to the raising numbers of images in the input set. To the best of our knowledge, this evolution ability has not been demonstrated by previous works before.

In this work, we use the semantic knowledge from CLIP to refine features from a standard backbone network (\eg, ResNet50) for segmentation. Along the network top-down path, as features are diffused from low-level cues (\eg, edges) to high-level semantics \cite{han2017robust, zhang2018exfuse, li2019group}, the refinement is performed in a coarse-to-fine manner, acting on three intermediate feature maps from the backbone segmentation network. 

First, at the coarse level, we focus on encoding global semantic information with rich spatial details of the image set to coarse-level features. 
This is done in a graph message-passing framework by using all features from the image set to capture its common semantics and update one specific image feature. In this way, we achieve the goal of injecting global consistent semantics of the image set into features.

Second, at the middle level, we are ready to modulate backbone features by using CLIP, dubbed as CLIP interaction. Specifically, the same image set is fed to the image encoder of CLIP to extract discriminative image embeddings. After fusing image embeddings with pre-defined template text embeddings, we obtain semantic embeddings from CLIP and use the semantics to refine middle-level features.

Finally, note that the pre-trained CLIP is general, and feature embeddings from it do not necessarily focus only on the common objects. To draw CLIP towards this co-segmentation task, we propose to use a small multi-layer perceptron network to identify the most useful CLIP embedding. We use it to refine the finest backbone feature similarly, regularizing the backbone feature towards the most common semantic class of the image set. This step is dubbed as CLIP regularization.

Experimentally, we demonstrate state-of-the-art performance on standard benchmarks.

Our codes and models will be released to facilitate reproducible research. 

To summarize, our contributions are given below,

\begin{itemize}
\vspace{-.5em}
    \item We propose a framework for leveraging CLIP for the co-segmentation task.
    \item We design an image set feature correspondence module to encode the global semantics of the image set.
    \item We design CLIP interaction and regularization modules to mine common semantics in a coarse-to-fine manner.
    \item We draw CLIP towards the co-segmentation task by using a small multi-layer perceptron network, which is optimized by a carefully tailored classification loss.
\end{itemize}

\section{Related Work}
\paragraph{Co-Segmentation.} The key difficulty of co-segmentation tasks is extracting common semantics from an image set \cite{liang2017multi}. Existing methods can be categorized into pair-wise correlation, multi-task, and iteration based models. Pair-wise correlation models employ siamese networks to extract common semantics of each image pair \cite{chen2018semantic, li2018deep}, yet their results are often sub-optimal due to semantic ambiguity existing in the whole image set. Multi-task-based models attempt to address the ambiguity by explicitly constraining the network on common semantic class classification \cite{zhang2020deep, su2022unified}, requiring extra manual annotation from training data. Meanwhile, iteration based models propose to resolve the common semantic ambiguity by recurrently reasoning common semantics and refining predictions \cite{li2019group, zhang2021cyclesegnet} which are computationally intensive. In contrast, we leverage the pre-trained CLIP model to effectively and efficiently reason common semantics in a single forward pass, without requiring extra semantic class annotation and an expensive recurrent refinement strategy.

\paragraph{Image Segmentation with CLIP.} The CLIP \cite{radford2021learning} performs contrastive learning on large-scale web-curated image-text pairs, showing promising zero-short learning capability. 
Existing methods extend the zero-shot classification ability to dense predictions by mainly following proposal classification or pixel classification based methods.
Proposal classification based methods introduce a mask proposal generator and uses CLIP to classify each masked image to ensemble the segmentation results \cite{xu2021simple, ding2022decoupling}. 
Pixel classification based methods generally employ CLIP as a pre-trained encoder and train a decoder to classify each pixel from CLIP features \cite{zhou2021denseclip, zhou2023zegclip}. 
However, both the methods require prior knowledge of ground-truth class semantics to perform segmentation. Our method distills common semantics from an image set, without the ground-truth common semantics.

\paragraph{Foundation Segmentation Models.} Pioneer works of foundation segmentation models can be found as SAM \cite{kirillov2023segany} and SEEM \cite{zou2023segment}. They frame the zero-shot segmentation into a universal promptable and interactive interface, taking points, bounding box, or semantic class of interests as inputs, and predicting the refereed segmentation masks. In this paper, we compare with these foundation models on the co-segmentation task. Though providing them with ground-truth bounding boxes and semantic classes of the common semantics, unsatisfactory segmentation results are generally obtained, indicating the necessity of in-depth studying for the co-segmentation problem.

\begin{figure*}[!t]
    \centering
    \includegraphics[width=.9\linewidth]{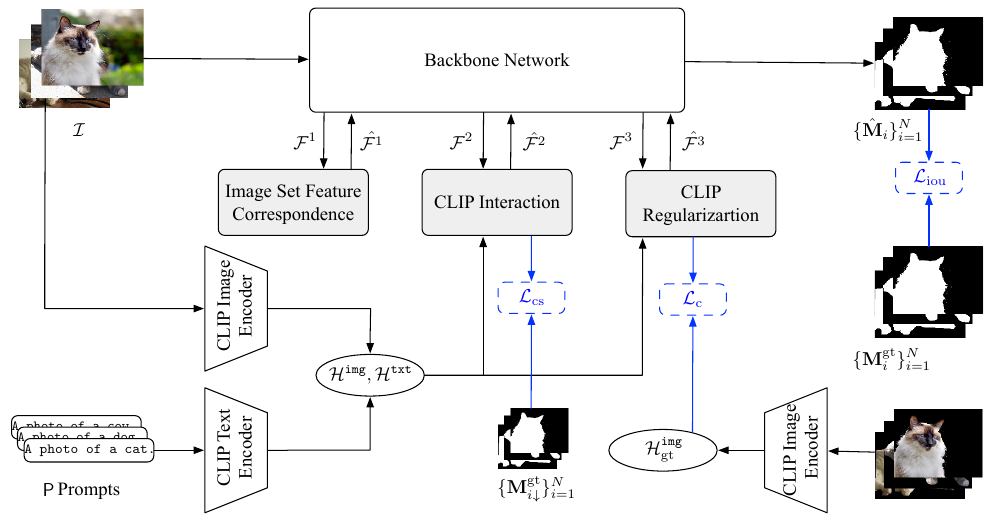}
    \caption{\it \small The architecture of our method. Given a set of images $\ \mathcal{I} = \{\mathbf{I}_{i}\}^{N}_{i=1}$, for each image, we aim to estimate a mask $\hat{\mathbf{M}}_{i}$ to segment their common semantic object with a backbone network and three key modules. 
    We feed each image individually to the backbone network, and use the three modules to refine 
    intermediate backbone features in a coarse-to-fine manner. In the image set feature correspondence module, we encode global consistent semantic information within images to {refine} feature $\mathcal{F}^{1}$. In the CLIP interaction module, we use CLIP {embeddings} 
    $\mathcal{H}^{\tt{img}}$ and $\mathcal{H}^{\tt{txt}}$ to refine feature $\mathcal{F}^{2}$. In the CLIP regularization module, we use CLIP embeddings to mine the most common semantic within the image set, and use the semantic to regularize the backbone feature $\mathcal{F}^{3}$. 
    While keeping the CLIP model frozen, our model is optimized with three losses. A $\mathcal{L}_{\text{iou}}$ to encourage the predict masks $\{\hat{\mathbf{M}}_{i}\}_{i=1}^{N}$ overlapping with the ground-truth masks $\{\mathbf{M}^{\text{\rm gt}}_{i}\}_{i=1}^{N}$.
    A coarse segmentation loss $\mathcal{L}_{\text{cs}}$ to optimize the CLIP interaction module, using downsampled ground-truth masks $\{\mathbf{M}^{\text{\rm gt}}_{i \downarrow}\}_{i=1}^{N}$. A classification loss $\mathcal{L}_{\text{c}}$ to optimize the CLIP regularization module, 
    using CLIP embeddings extracted from ground-truth masked images.}
    \label{fig:overview}
    \vspace{-1.5em}
\end{figure*}

\section{Methods}

\paragraph{Problem Formulation.} Given a set of images $\mathcal{I} = \{\mathbf{I}_{i}\}_{i=1}^{{N}}$ containing a common semantic object, for each image, we aim to estimate the mask $\hat{\mathbf{M}}_{i}$ of the object, where ${N}$ is the number of images. Here, $\mathbf{I}_{i} \in \mathbb{R}^{\mathsf{H} \times \mathsf{W} \times 3}$, $\hat{\mathbf{M}}_{i} \in \mathbb{R}^{\mathsf{H} \times \mathsf{W} \times 1}$, $\mathsf{H}$ and $\mathsf{W}$ are the height and width of an image.

\paragraph{Overview.} Our main idea is using CLIP to refine multi-scale intermediate features from a backbone network $f(\cdot)$, which takes an image as an input and estimates a mask. 

Feeding each image from the set $\mathcal{I}$ to $f(\cdot)$, we collect three coarse-to-fine intermediate features, which are denoted as $\mathcal{F}^{1} = \{\mathbf{F}^{1}_{i}\}_{i=1}^{{N}}$, $\mathcal{F}^{2} = \{\mathbf{F}^{2}_{i}\}_{i=1}^{{N}}$, and $\mathcal{F}^{3} = \{\mathbf{F}^{3}_{i}\}_{i=1}^{{N}}$. 
At the same time, with a pre-trained CLIP, we feed $\forall \mathbf{I}_{i} \in \mathcal{I}$ to the CLIP image encoder to get image embeddings $\mathcal{H}^{\tt{img}} = \{\mathbf{h}^{\tt{img}}_{i}\}_{i=1}^{{N}}$, and CLIP text embeddings $\mathcal{H}^{\tt{txt}} = \{\mathbf{h}^{\tt{txt}}_{i}\}_{i=1}^{{P}}$ by feeding $P$ pre-defined prompts to the CLIP text encoder. 

With $\mathcal{H}^{\tt{img}}$ and $\mathcal{H}^{\tt{txt}}$, we are ready to refine intermediate features $\mathcal{F}^{1}$, $\mathcal{F}^{2}$ and $\mathcal{F}^{3}$ in a coarse-to-fine manner:
i) an image set feature correspondence module to encode global consistent semantic information within $\mathcal{F}^{1}$;
ii) a CLIP interaction module to refine $\mathcal{F}^{2}$ based on the CLIP 

embeddings  
$\mathcal{H}^{\tt{img}}$ and $\mathcal{H}^{\tt{txt}}$;
iii) a CLIP regularization module to regularize the semantic of $\mathcal{F}^{3}$ towards the most common semantic class of $\mathcal{I}$. 

A segmentation loss and classification loss are proposed for the CLIP interaction and regularization modules, respectively. The architecture of our method is given in Fig.~\ref{fig:overview}.

\subsection{Image Set Feature Correspondence}
\label{sec:image_set}
We first inject global consistent semantics 
of the image set into each feature map $\mathbf{F}^{1}_{i} \in \mathcal{F}^{1}$. We aim to make $\mathbf{F}^{1}_{i}$ focus on the common object within the image set. We drop the superscript of $\mathbf{F}^{1}_{i}$, for clarity.

Inspired by the success of the attention mechanism and graph neural network, we use cross-attention to aggregate global image set information. We first define a complete graph for the image set, denoted by $\mathcal{G}$. Nodes in $\mathcal{G}$ correspond to images, node values correspond to image features $\mathbf{F}_{i}$, and edges connect all images. 

Node values are updated using multi-head cross-attention in a message-passing framework \cite{hamilton2017inductive, sarlin2020superglue}. For an edge connecting the $i^\text{th}$ and $j^\text{th}$ nodes ($i \neq j$), the node value $\mathbf{F}_{i}$ is updated to $\bar{\mathbf{F}}^{j}_{i}$ via
\begin{equation}
    \bar{\mathbf{F}}^{j}_{i} = {\mathbf{F}}_{i} + \text{FFN}_1\biggl( {\mathbf{F}}_{i} || 
    \mathfrak{m}_{\mathbf{F}_{j} \to \mathbf{F}_{i}} \biggr) \ ,
\end{equation}
where $\cdot||\cdot$  denotes  concatenation, $\mathfrak{m}_{\mathbf{F}_{j} \to \mathbf{F}_{i}}$ denotes message from
node $j$ to node $i$, and $\text{FFN}_1(\cdot)$ is a 
convolutional feed-forward network. The message $\mathfrak{m}_{\mathbf{F}_{j} \to \mathbf{F}_{i}}$ is calculated through the standard attention \cite{vaswani2017attention} via $\text{Att}({\mathbf{F}}_{i}, {\mathbf{F}}_{j}, {\mathbf{F}}_{j})$, with query ${\mathbf{F}}_{i}$ and key/value ${\mathbf{F}}_{j}$.

For each edge connecting node $i$, we can compute its updated node value. For node $i$, collecting all updated node values results in a set $\{\bar{\mathbf{F}}^{j}_{i} | j = 1,...,N, j\neq i\}$. We compute a weight $\boldsymbol{\alpha}^{j}_{i}$ of each $\bar{\mathbf{F}}^{j}_{i}$, and perform a weighted average of the set to compute the final node value update. 
To compute $\boldsymbol{\alpha}^{j}_{i}$, we stack the set along the feature channel dimension and perform a softmax normalization $\text{Softmax}(\cdot)$ along feature channels. After splitting the stacked tensor, we obtain $\boldsymbol{\alpha}^{j}_{i}$. The final node value update $\hat{\mathbf{F}}_{i}$ is given by 
\begin{equation}
    \hat{\mathbf{F}}_{i} = \text{Conv} \biggl ( \sum_{j=1, j\neq i}^{N} \bigl ( \boldsymbol{\alpha}^{j}_{i} \odot \bar{\mathbf{F}}^{j}_{i} \bigr ) \biggr ) \ ,
\end{equation}
where $\odot$ denotes Hadamard (element-wise) product, and $\text{Conv}(\cdot)$ is a simple convolution layer.

\subsection{CLIP Interaction}
\label{sec:clip_interaction}
 
Given $\mathcal{I} = \{\mathbf{I}_{i}\}_{i=1}^{{N}}$, we use CLIP to mine accurate common semantics within the image set, and inject the semantics into $\mathbf{F}^{2}_{i} \in \mathcal{F}^{2}$ to refine the feature map. In the following, we first briefly summarize CLIP for self-contain purposes, then compute semantic embeddings  
with CLIP, and finally use the semantic embeddings 
to refine $\mathbf{F}^{2}_{i}$.  

\paragraph{CLIP Preliminary.} The CLIP separately embeds an image and a paired text description with an image encoder and a text encoder into the same feature space. The CLIP optimizes a contrastive loss to pull embeddings of aligned images and texts close to each other, while pushing away embeddings of misaligned pairs. By training on 400 million text-image pairs, CLIP shows promising zero-shot learning performance that aligns images with prompts of open-world descriptions. For a pair of image and text, the similarity between the image embedding $\mathbf{h}^{\tt{img}}$ and text embedding $\mathbf{h}^{\tt{txt}}$ is large if they are aligned, and small if misaligned.

\begin{figure}
\begin{center}
    
    \setlength{\tabcolsep}{3pt}
    \begin{tabular}{cccc}
      \includegraphics[width=.225\linewidth]{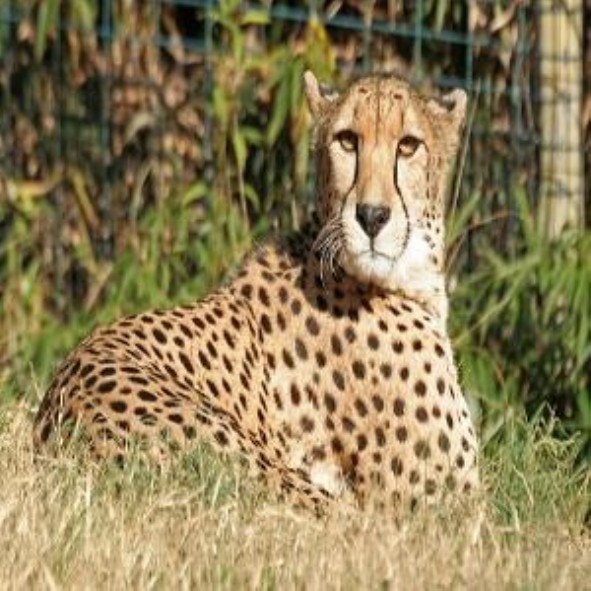}  & \includegraphics[width=.225\linewidth]{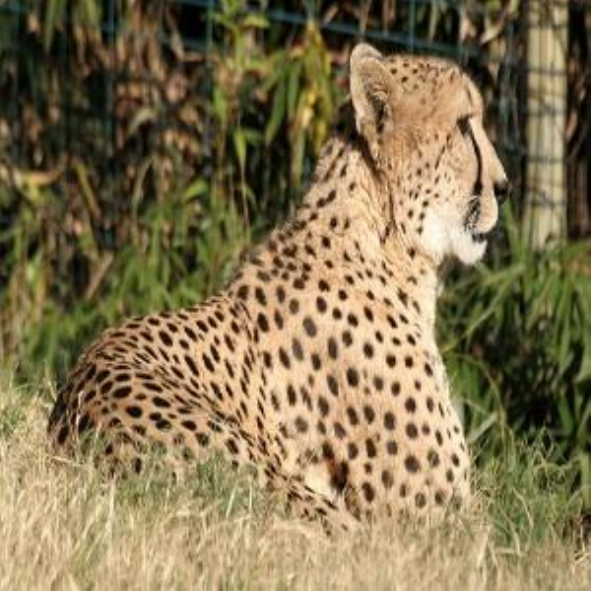}  & \includegraphics[width=.225\linewidth]{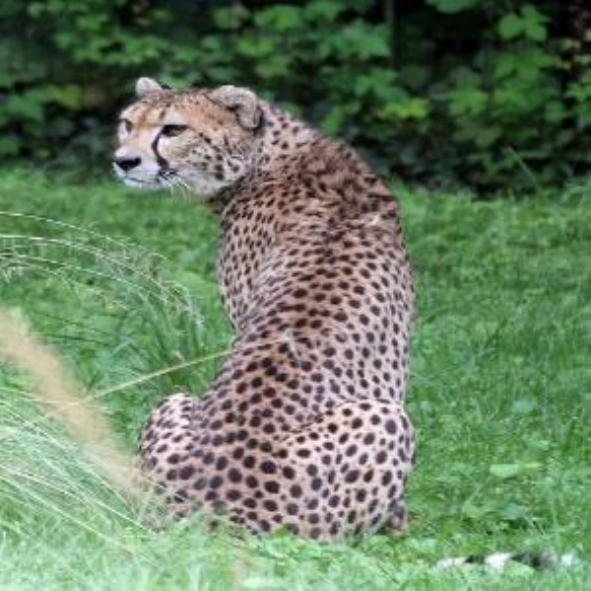}  & \includegraphics[width=.225\linewidth]{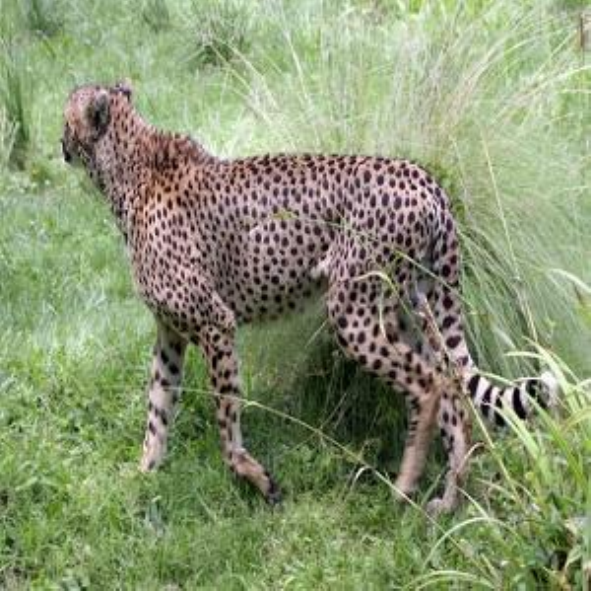}  \\
      \includegraphics[width=.225\linewidth]{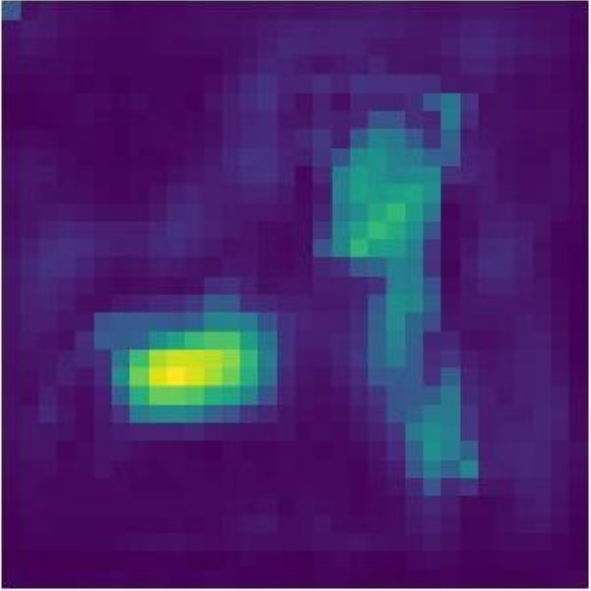}  & \includegraphics[width=.225\linewidth]{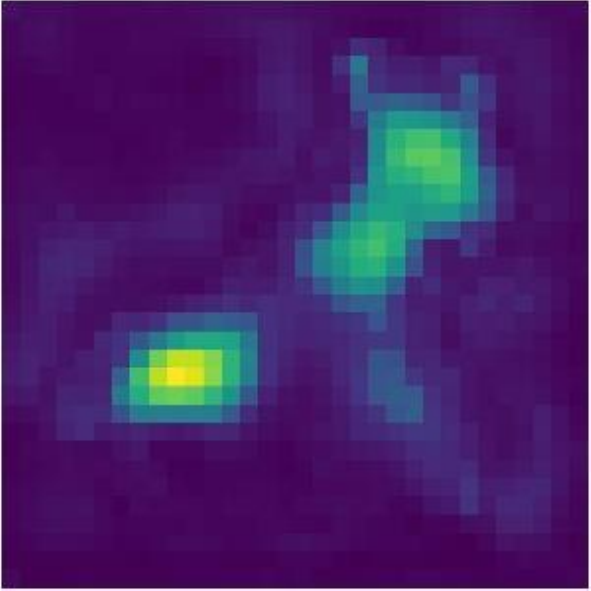}  & \includegraphics[width=.225\linewidth]{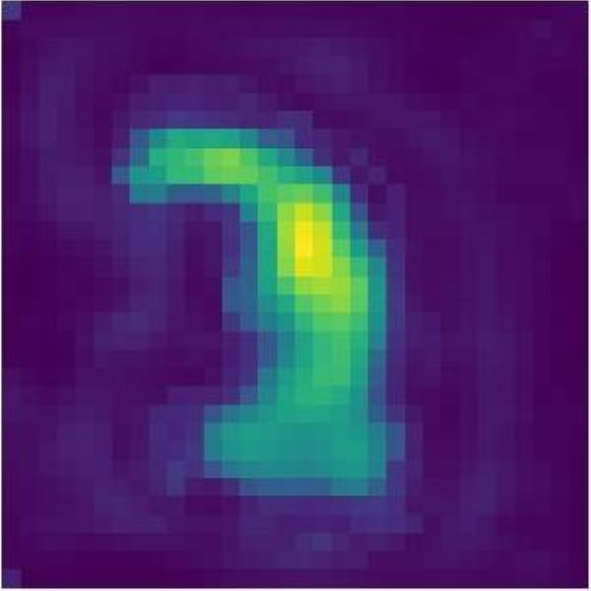}  & \includegraphics[width=.225\linewidth]{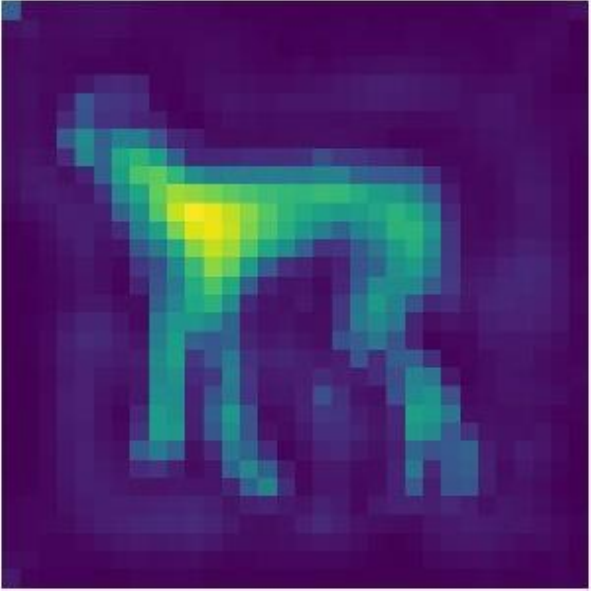}  \\
      \includegraphics[width=.225\linewidth]{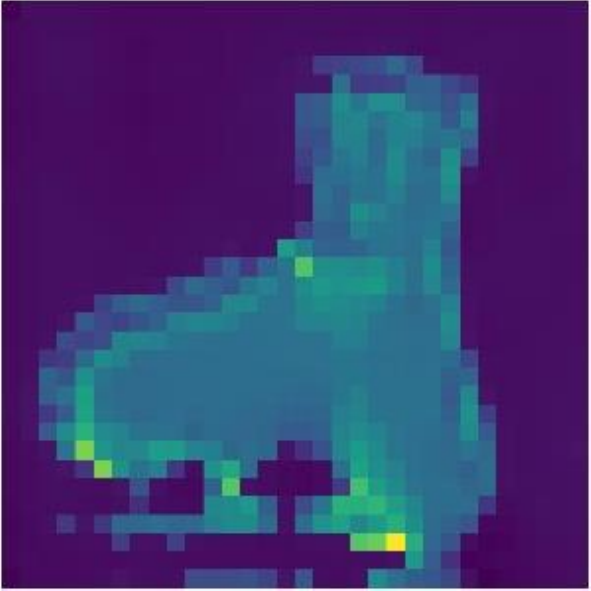}  & \includegraphics[width=.225\linewidth]{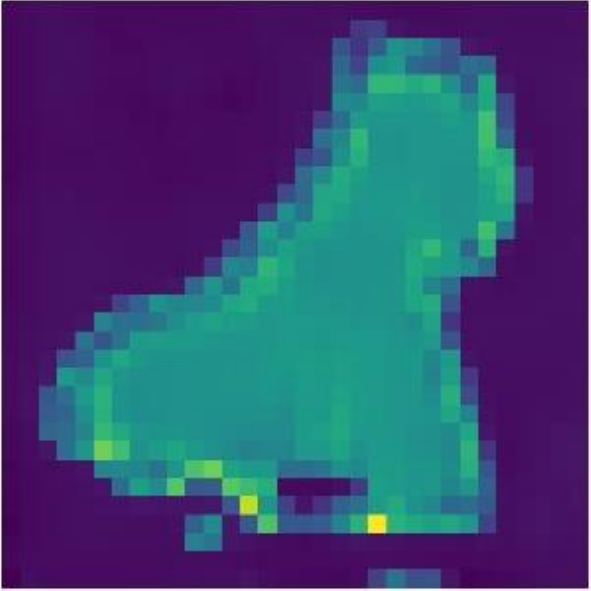}  & \includegraphics[width=.225\linewidth]{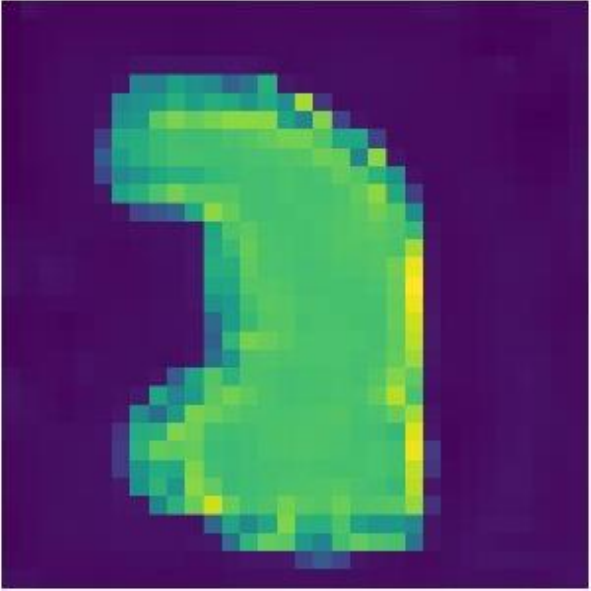}  & \includegraphics[width=.225\linewidth]{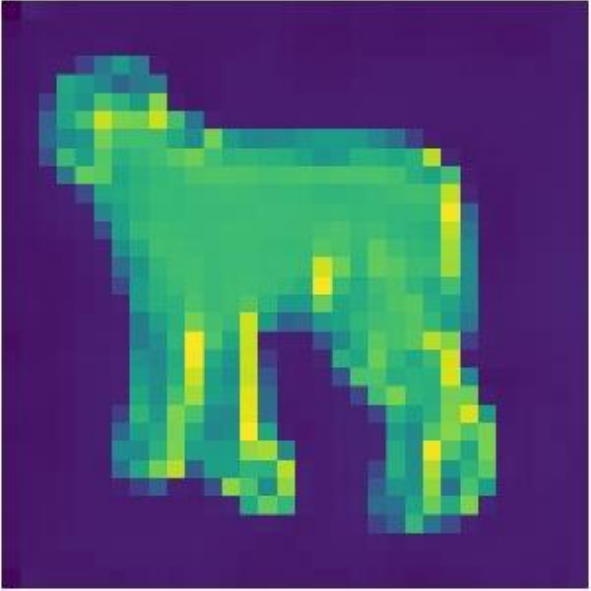}  \\
    \end{tabular}
   
    \setlength{\tabcolsep}{2.5pt}
    \begin{tabular}{ccccc}
    \toprule
     Dataset  &  MSRC  & Internet & iCoseg & PASCAL\\
    \hline
    {\small $\|\mathbf{M}_i^\text{gt} \odot \text{Softmax}(\mathbf{F}^{3}_{i}) \|$} & 0.035 & 0.029 & 0.037 & 0.028 \\
    {\small $\|\mathbf{M}_i^\text{gt} \odot \text{Softmax}(\hat{\mathbf{F}}^{3}_{i}) \|$}  & 0.486  & 0.489 & 0.513 & 0.413 \\ 
    \bottomrule
    \end{tabular}
   
\end{center}
 \caption{\it \small (Top) Feature visualizations of our CLIP regularization module. We show a set of images in the $\mathit{1^{st}}$ row, and feature map before and after using our CLIP regularization module in the $\mathit{2^{ed}}$ and $\mathit{3^{rd}}$ rows, \ie, $\mathbf{F}^{3}_{i}$ and $\hat{\mathbf{F}}^{3}_{i}$ in Eq.~\eqref{eq:f3}.
 (Bottom) The quantitative comparison of the quality of the feature map $\mathbf{F}^{3}_{i}$ and $\hat{\mathbf{F}}^{3}_{i}$. We calculate the standard $l^2$ norm of the feature map by using ground-truth co-segmentation masks $\mathbf{M}_i^\text{\rm gt}$, \eg, $\|\mathbf{M}_i^\text{\rm gt} \odot \text{Softmax}(\mathbf{F}^{3}_{i}) \|$, on the four well-known co-segmentation datasets (the higher the better). 
 }
\label{fig:visualization}
\vspace{-1.5em}
\end{figure}

\paragraph{Text Semantic Distillation.} 
We have a set of CLIP text embeddings $\mathcal{H}^{\tt{txt}} = \{\mathbf{h}^{\tt{txt}}_{i}\}_{i=1}^{{P}}$, obtained by feeding $P$ prompts to the text encoder of CLIP. Each prompt describes a potential semantic class of an image, \eg, \texttt{A photo of a [CLASS]}. Note that $\mathcal{H}^{\tt{txt}}$ is independent of images $\mathcal{I}$, fixed, and complete, \ie, combining semantics contained in the set leads to the semantic of a novel class.

We then feed each image in $\mathcal{I}$ to the image encoder of CLIP, and obtain the CLIP image embedding $\mathcal{H}^{\tt{img}} = \{\mathbf{h}^{\tt{img}}_{i}\}_{i=1}^{{N}}$. To distill aligned text embeddings with $\mathcal{I}$ from $\mathcal{H}^{\tt{txt}}$, we compute the pairwise cosine similarity between feature embeddings in  $\mathcal{H}^{\tt{img}}$ and $\mathcal{H}^{\tt{txt}}$, and obtain a similarity matrix $\mathbf{S} \in \mathbb{R}^{N \times P}$. Collecting all similarities from $\mathcal{H}^{\tt{img}}$ by summarizing rows of $\mathbf{S}$, we obtain a similarity vector $\boldsymbol{\sigma} \in \mathbb{R}^{1 \times P}$. The element ${\sigma}_i$ describes the alignment/matchedness of text embedding $\mathbf{h}^{\tt{txt}}_{i}$ with respect to $\mathcal{I}$. By finding the Top-K elements in $\boldsymbol{\sigma}$, we distill a subset of CLIP text embeddings that align with $\mathcal{I}$. The distilled CLIP text embedding set is denoted by $\mathcal{H}^{\tt{txt}}_{\mathcal{K}} = \{\mathbf{h}^{\tt{txt}}_{i} \mid \sigma_i \in \text{TopK}(\boldsymbol{\sigma}) \}$.

\paragraph{Text-Image Semantic Fusion.} Given CLIP image embedding $\mathcal{H}^{\tt{img}}$ and distilled text embeddings $\mathcal{H}^{\tt{txt}}_{\mathcal{K}}$, we fuse the two semantics 
to compute a single CLIP semantic.

Note that image embedding $\mathbf{h}^{\tt{img}}_{i}$ is obtained independently for each $\mathbf{I}_i$. To impose global consistent semantic constraint, 
we inject global CLIP image semantic 
to each image embedding $\mathbf{h}^{\tt{img}}_{i}$, and obtain refined image embedding $\hat{\mathbf{h}}^{\tt{img}}_{i}$,
\begin{equation}\label{eq::MLP1}
    \hat{\mathbf{h}}^{\tt{img}}_{i} = \text{MLP}_{1}\biggl(\text{CAT}\bigl[\mathbf{h}^{\tt{img}}_{i}, \mathbf{\bar{h}}^{\tt{img}}\bigr]\biggr) \ ,
\end{equation}
where $\text{CAT}[\cdot,\cdot]$ denotes concatenation along the feature dimension, and $\text{MLP}_{1}(\cdot)$ is a small multi-layer perceptron. $\mathbf{\bar{h}}^{\tt{img}} = \frac{1}{{N}}\sum^{N}_{i=1} \mathbf{h}^{\tt{img}}_{i}$ is the global CLIP image semantic.

After imposing global semantic consistency constraint, we fuse text and image semantics by 
\begin{equation}
    \mathbf{z}^{\tt{img}}_{i} = \hat{\mathbf{h}}^{\tt{img}}_{i} + \text{MLP}_{2}\Bigl( 
    \mathfrak{m}_{[\hat{\mathcal{H}}^{\tt{img}},  \mathcal{H}^{\tt{txt}}_{\mathcal{K}}] \to \hat{\mathbf{h}}^{\tt{img}}_{i}} \Bigr) \ ,
\end{equation}
where $\text{MLP}_{2}(\cdot)$ denotes a multi-layer perception, and $\mathfrak{m}_{[\hat{\mathcal{H}}^{\tt{img}},  \mathcal{H}^{\tt{txt}}_{\mathcal{K}}] \to \hat{\mathbf{h}}^{\tt{img}}_{i}}$ is message from all CLIP image embeddings and distilled text embeddings to $\hat{\mathbf{h}}^{\tt{img}}_{i}$. The message is calculated through the standard attention \cite{vaswani2017attention} via $\text{Att}({\hat{\mathbf{h}}^{\tt{img}}_{i}}, \mathcal{H}^{\tt{it}}, \mathcal{H}^{\tt{it}})$, with query $\hat{\mathbf{h}}^{\tt{img}}_{i}$ and key/value $\mathcal{H}^{\tt{it}}$, where $\mathcal{H}^{\tt{it}} = \text{STK}[\hat{\mathcal{H}}^{\tt{img}},  \mathcal{H}^{\tt{txt}}_{\mathcal{K}}]$ and $\text{STK}[\cdot,\cdot]$ denotes stack along the length dimension to generate $N+K$ embeddings.

\paragraph{Semantic Modulation.} With CLIP semantic feature $\mathbf{z}^{\tt{img}}_{i}$, we are ready to refine the semantic feature map $\mathbf{F}^{2}_{i} \in \mathcal{F}^{2}$,
\begin{equation}
        \hat{\mathbf{F}}^{2}_{i} = \text{FFN}_{\text{3}} \biggl(\text{FFN}_{2}\bigl(\mathbf{F}^{2}_{i}\bigr) \odot \text{PAD}\bigl(\mathbf{z}^{\tt{img}}_{i}\bigr)\biggr) \ , \label{eq:modul}
\end{equation}
where the convolutional feed-forward network $\text{FFN}_{2}(\cdot)$ project $\mathbf{F}^{2}_{i}$ to the same embedding space of $\mathbf{z}^{\tt{img}}_{i}$. $\text{PAD}(\cdot)$ pads the embedding $\mathbf{z}^{\tt{img}}_{i}$ to the same spatial size as $\mathbf{F}^{2}_{i}$. $\text{FFN}_{3}(\cdot)$ projects refined feature map to the  same embedding space of $\mathbf{F}^{2}_{i}$.

\begin{table*}[htbp]

\caption{\it \small Comparison with respect to state-of-the-art methods on the MSRC, Internet, iCoseg, and PASCAL dataset under different training datasets. Note, Zhang \cite{zhang2020deep} and Su \cite{su2022unified} use the ground-truth class labels in their training phase. The best results are in {\textbf{bold}}. }

\centering
\setlength{\tabcolsep}{5pt}

\begin{tabular}{lccccccccc}
\toprule
\multirow{2}*{Method} & \multirow{2}*{Train} & \multicolumn{2}{c}{MSRC} & \multicolumn{2}{c}{Internet} & \multicolumn{2}{c}{iCoseg} & \multicolumn{2}{c}{PASCAL}\\
\cmidrule{3-10}
~ & ~ &  \multicolumn{1}{c}{$\mathcal{P}$ (\%)} &  \multicolumn{1}{c}{$\mathcal{J}$ (\%)} & \multicolumn{1}{c}{$\mathcal{P}$ (\%)} &  \multicolumn{1}{c}{$\mathcal{J}$ (\%)} & \multicolumn{1}{c}{$\mathcal{P}$ (\%)} &  \multicolumn{1}{c}{$\mathcal{J}$ (\%)}  & \multicolumn{1}{c}{$\mathcal{P}$ (\%)} &  \multicolumn{1}{c}{$\mathcal{J}$ (\%)} \\ 
\midrule
  Vicente  \cite{vicente2011object}   & -  &   90.2 &  70.6 & - & - & - & - & - & - \\
Wang \cite{wang2013image}  & -        &  92.2  & - & - & - & - & - & - & -\\
Rubinstein \cite{rubinstein2013unsupervised}  & -  &  92.2   & 74.7 & 85.4 & 57.6 & - & 70.2 & - & -\\
  Faktor \cite{faktor2013co}                 & -   &  92.0 &  77.0 & - & - & 92.8 & 73.8 & - & - \\
 Quan \cite{quan2016object} & - & - & - & 89.6 & 60.4 & 94.8 & 82.0 & 89.0 & 52.0 \\
Jerripothula \cite{jerripothula2016image}    &  - & 88.7 & 71.0 & 88.9 & 64.0 & 91.9 & 72.0 & 85.2 & 45.0 \\
Wang \cite{Wang2017Mu} & - & 90.9 & 73.0 & - & - & 93.8 & 77.0 & 84.3 & 52.2  \\ 
Yuan   \cite{yuan2017deep}  & PASCAL & - & - & 91.1 & 67.7 & 96.0 & 86.0 & - & - \\
Chen \cite{chen2018semantic}   & PASCAL   &  95.3   &   77.7  & - & 73.1 & - & 86.0 & - & 59.8\\
Li \cite{li2018deep} & PASCAL           &  95.4  & 82.9   & 93.5 & 72.6 & - & 84.2 & 94.2 & 64.5\\
Zhang \cite{zhang2021cyclesegnet}    & PASCAL    & \textbf{97.9} & 87.2 & - & 80.4 & - & 90.8 & 95.8 & 75.4\\
\textbf{Ours}  & PASCAL &  97.0 & \textbf{88.4} & \textbf{95.4} & \textbf{82.1} & \textbf{97.8} & \textbf{91.7} & \textbf{96.1} & \textbf{75.9} \\ \midrule

 Li  \cite{li2019group} & COCO & - & - & 97.1 & 84.0 & 97.9 & 89.0 & 94.1 & 63.0\\
Zhang \cite{zhang2020deep}  & COCO     &   95.2 & 81.9  & 93.6 & 74.1 & - & 89.2 & 94.9 & 71.0\\
Zhang \cite{zhang2021cyclesegnet}  & COCO    &   97.6  & 89.6  & - & 86.2 & - & 92.1 & 96.8 & 73.6\\
Su \cite{su2022unified} & COCO &  97.8 &  84.3 & 95.2 & 74.6 & 98.1 & 92.3 & 96.9 & 75.7\\

\textbf{Ours}   & COCO & \textbf{97.9}  & \textbf{89.8} & \textbf{97.6} & \textbf{87.5} & \textbf{98.3} & \textbf{92.9} & \textbf{97.1} & \textbf{76.4}\\
\bottomrule
\end{tabular}
\label{icoseg}
\vspace{-1.5em}
\end{table*}

\subsection{CLIP Regularization}
\label{sec:clip_regularization}
We refine features $\mathcal{F}^{1}$, $\mathcal{F}^{2}$, and $\mathcal{F}^{3}$ in a coarse-to-fine manner. After refining coarse and middle-level features $\mathcal{F}^{1}$ and $\mathcal{F}^{2}$, the fine-grained feature $\mathcal{F}^{3}$ becomes discriminative, ready to be used to predict the most common object within the image set. To regularize $\mathcal{F}^{3}$, we identify the most likely class from CLIP, and use its semantics to refine $\mathcal{F}^{3}$.

We have computed the CLIP image-to-text similarity matrix $\mathbf{S} \in \mathbb{R}^{N \times P}$ in {\emph{Sec. Clip Interaction}. To identify the most likely class within $P$ classes, we first split $\mathbf{S}$ into $N$ row vectors $\{\mathbf{s}_i | i=1,...,N\}$, with vector dimension of $P$. We then feed the $N$ vectors to a small MLP, followed by a global max pooling and $\text{Softmax}(\cdot)$, to estimate a similarity probability vector $\boldsymbol{\upsilon } \in \mathbb{R}^{1 \times P}$, and $\boldsymbol{\upsilon } = [\upsilon_1, \cdots, \upsilon_i, \cdots, \upsilon_P]$. By finding the largest similarity in $\boldsymbol{\upsilon }$, we obtain the most likely class $i^{*}$. Mathematically, we have,
\begin{align}
i^{*} &= \mathop{\arg \max}\limits_{\small i \in [1,P]}  \boldsymbol{\upsilon } \ , \\
\boldsymbol{\upsilon } &= \text{Softmax}\biggl(\text{MAX}\bigl( \text{MLP}_3(\{\mathbf{s}_{i}\} )\bigr)\biggr) \ ,
    \label{eq:pscla}
\end{align}
where $\text{MAX}(\cdot)$ denotes global max pooling. We use the CLIP embedding corresponding to the most likely class $i^{*}$ to regularize semantic feature map $\mathbf{F}^{3}_{i} \in \mathcal{F}^{3}$,
\begin{equation} \label{eq:f3}
        \hat{\mathbf{F}}^{3}_{i} = \text{FFN}_{\text{5}} \biggl(\text{FFN}_{4}\bigl(\mathbf{F}^{3}_{i}\bigr) \odot \text{PAD}\bigl(\mathbf{h}^{\tt{txt}}_{i^{*}}\bigr)\biggr) \ ,
\end{equation}
where the definitions of $\text{FFN}_{\text{5}}(\cdot)$, $\text{FFN}_{\text{4}}(\cdot)$, and $\text{PAD}(\cdot)$ are similar to  Eq.~\eqref{eq:modul}. Sample visualizations of $\mathbf{F}^{3}_{i}$ and $\hat{\mathbf{F}}^{3}_{i}$ are given in Fig.~\ref{fig:visualization}.

\subsection{Network Training}
\label{sec:losse}
Our network is trained with an IoU loss, a coarse segmentation loss, and a classification loss. Our training loss $\mathcal{L}_{\text{total}}$ is given by,

\begin{equation}\label{eq::totalLoss}
    \mathcal{L}_{\text{total}} = \mathcal{L}_{\text{iou}} + \lambda_{1} \mathcal{L}_{\text{cs}} + \lambda_{2} \mathcal{L}_{\text{c}} \ ,
\end{equation}
where $\lambda_{1}$ and $\lambda_{2}$ are hyperparameters.

\paragraph{IoU Loss.} We encourage the predicted co-segmentation masks to overlap with the ground-truth co-segmentation masks, by averaging IoU losses $\eta(\cdot,\cdot)$ \cite{su2022unified}. The loss is given by, 
\begin{equation}
    \mathcal{L}_{\text{iou}} = \frac{1}{N}\sum_{i=1}^{N} \eta\bigl(\hat{\mathbf{M}}_{i}, \mathbf{M}_{i}^{\text{gt}}\bigr)  \ ,
\end{equation}

where $\hat{\mathbf{M}}_{i}$ and ${\mathbf{M}}_{i}^\text{gt}$ denote estimated co-segmentation mask and ground-truth mask of the $i^\text{th}$ image, respectively.

\paragraph{Coarse Segmentation Loss.}
To regularize $\text{MLP}_{1}(\cdot)$ (Eq.~\eqref{eq::MLP1}), we propose to use a light-weight decoder to estimate a coarse segmentation mask $\hat{\mathbf{M}}^{\text{c}}_{i} = \text{Decoder}(\hat{\mathbf{h}}^{\tt{img}}_{i})$. By minimizing the difference between $\hat{\mathbf{M}}^{\text{c}}_{i}$ and ground-truth, the $\text{MLP}_{1}(\cdot)$ is optimized. The loss is defined as,

\begin{align}
    \mathcal{L}_{\text{cs}} = \frac{1}{N}\sum_{i=1}^{N} \eta\bigl(\hat{\mathbf{M}}^{\text{c}}_{i}, \mathbf{M}_{i\downarrow}^\text{gt}\bigr) \ ,
\end{align}
where $\mathbf{M}_{i\downarrow}^\text{gt}$ is the downsampled ground-truth mask  of the $i^\text{th}$ image in the set.

\begin{figure*}
\begin{center}
    \begin{subtable}{\linewidth}
    \setlength{\tabcolsep}{3pt}
    \begin{tabular}{cccccccccccc}
      \multirow{2}*{Dog} &
      \includegraphics[width=.07\linewidth]{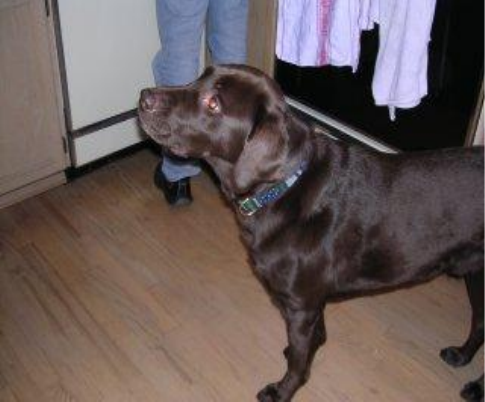}  & 
      \includegraphics[width=.07\linewidth]{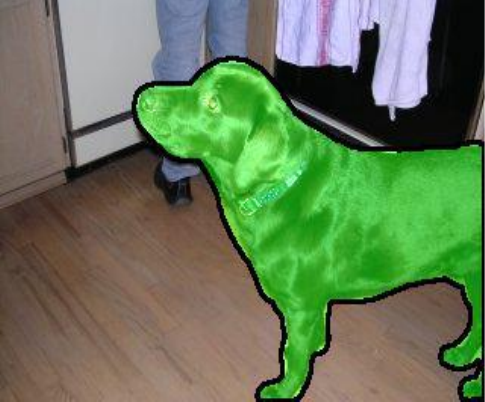}  & \includegraphics[width=.07\linewidth]{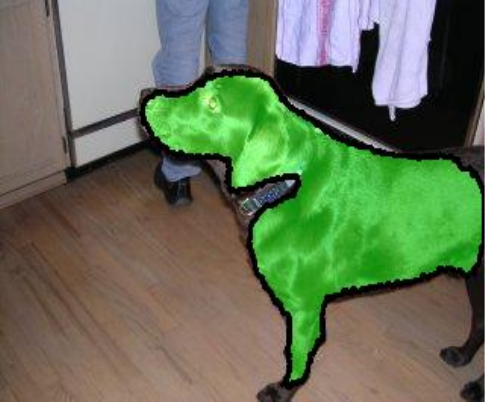}  & 
      \includegraphics[width=.07\linewidth]{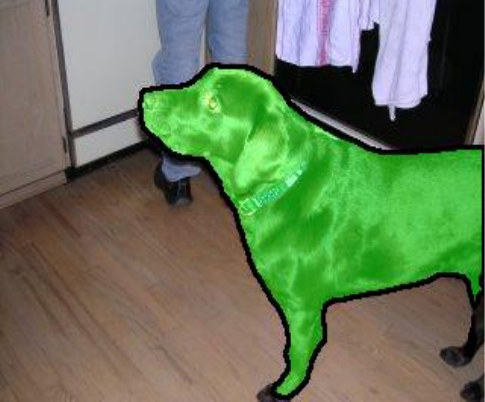}  & 
      \includegraphics[width=.07\linewidth]{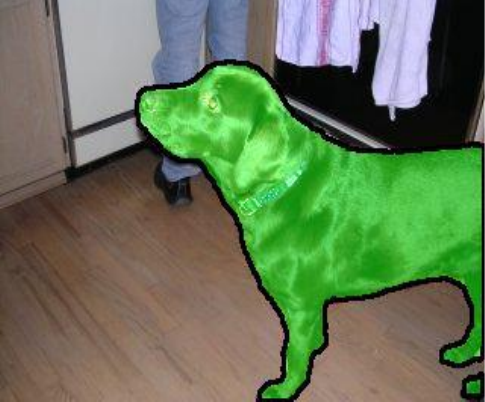} & 
      \multirow{2}*{Car} &
      \includegraphics[width=.07\linewidth]{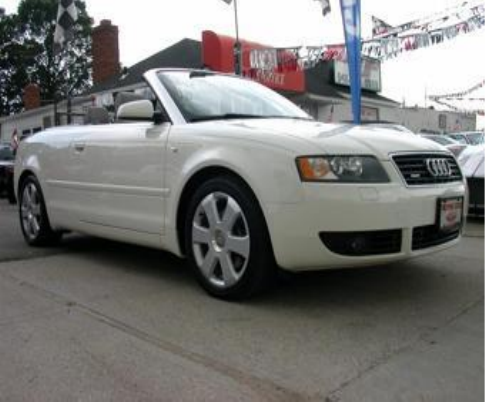} &
       \includegraphics[width=.07\linewidth]{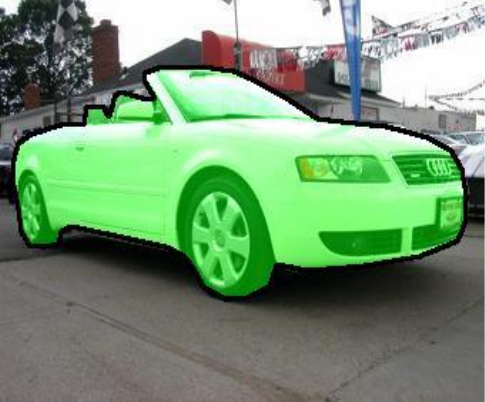} & \includegraphics[width=.07\linewidth]{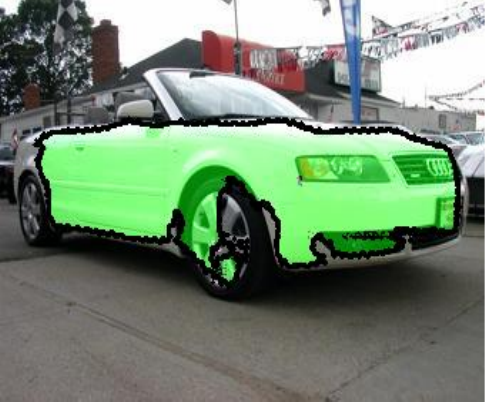} &
      \includegraphics[width=.07\linewidth]{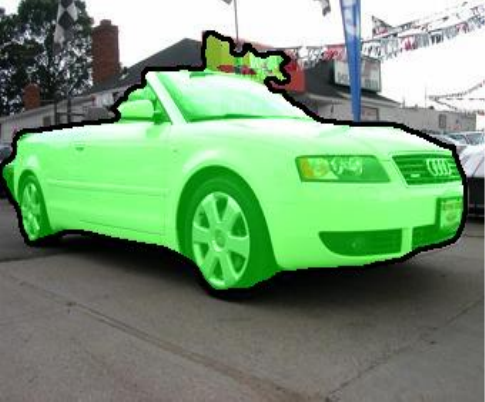} &
      \includegraphics[width=.07\linewidth]{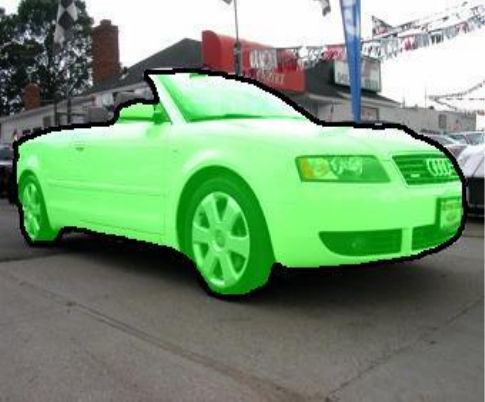} \\

      \multirow{2}*{} &
      \includegraphics[width=.07\linewidth]{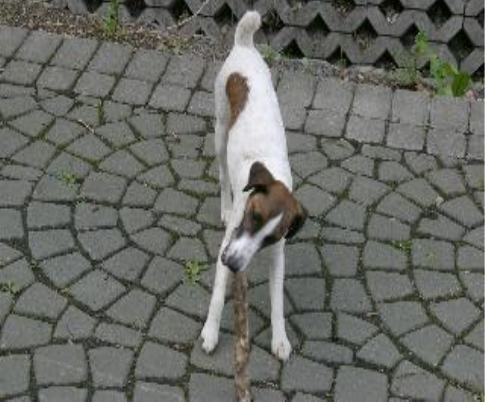}  & 
      \includegraphics[width=.07\linewidth]{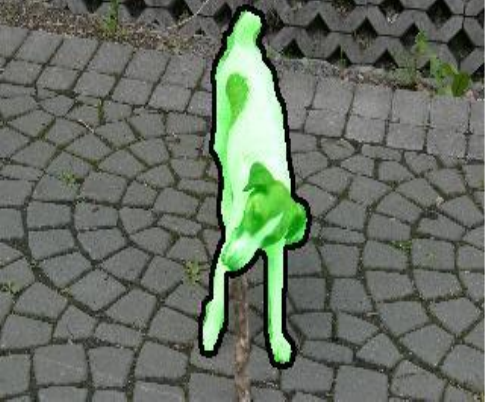}  & \includegraphics[width=.07\linewidth]{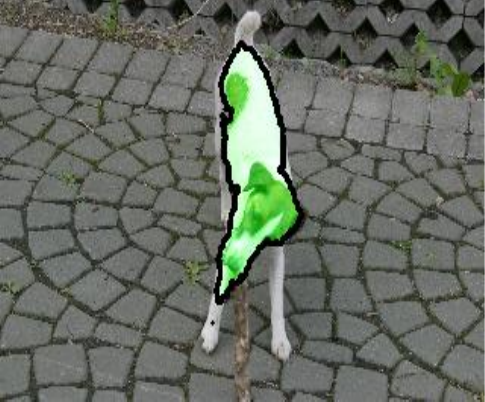}  & 
      \includegraphics[width=.07\linewidth]{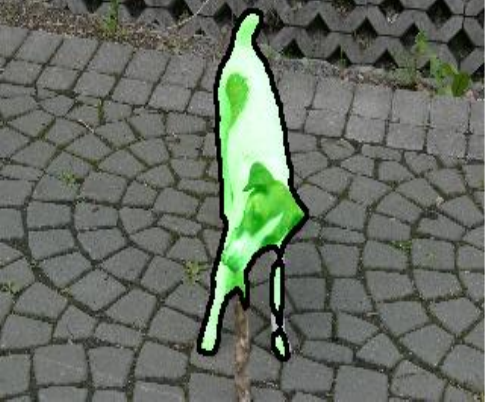}  & 
      \includegraphics[width=.07\linewidth]{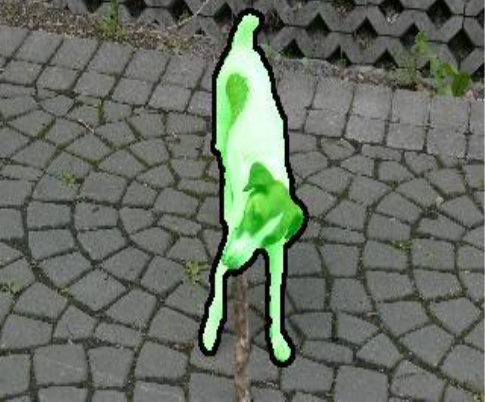} & 
      \multirow{2}*{} &
      \includegraphics[width=.07\linewidth]{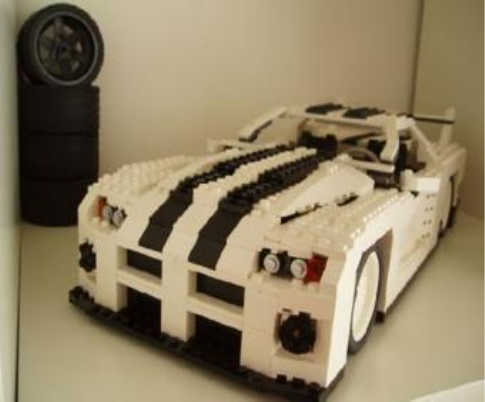} &
       \includegraphics[width=.07\linewidth]{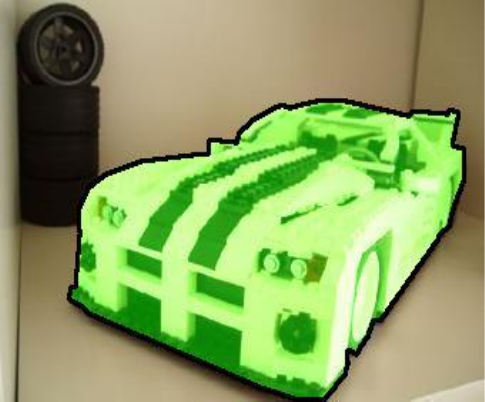} & \includegraphics[width=.07\linewidth]{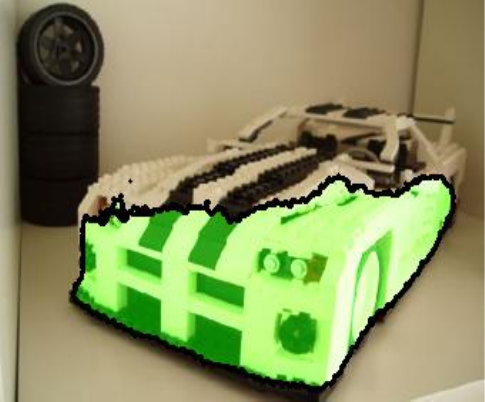} &
      \includegraphics[width=.07\linewidth]{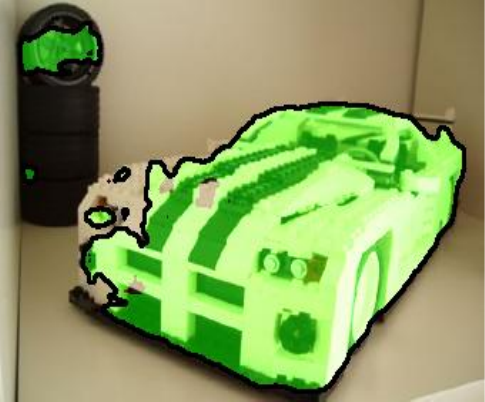} &
      \includegraphics[width=.07\linewidth]{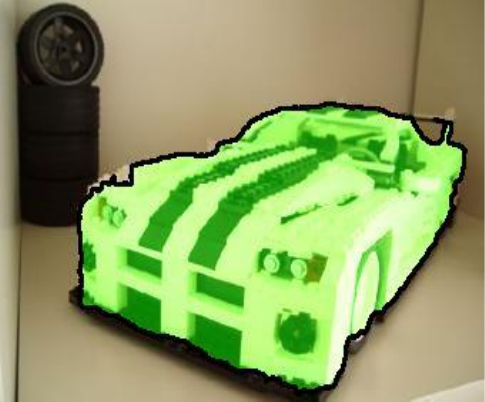} \\

      \multirow{2}*{Plane} &
      \includegraphics[width=.07\linewidth]{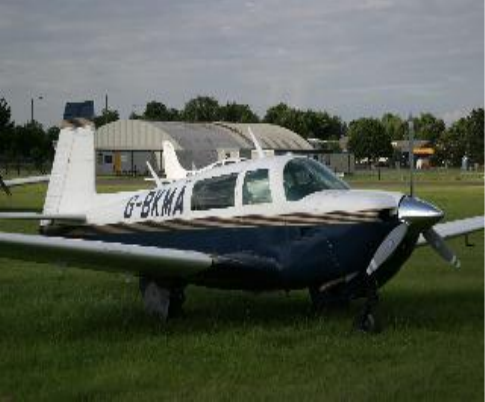}  & 
      \includegraphics[width=.07\linewidth]{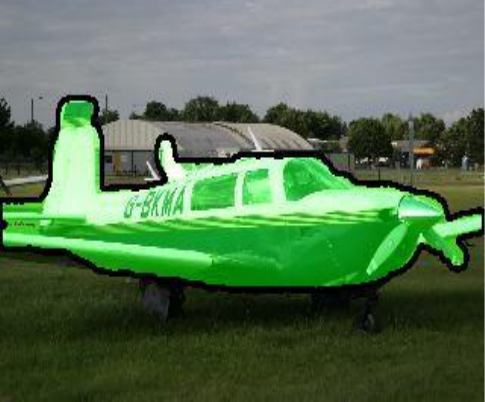}  & \includegraphics[width=.07\linewidth]{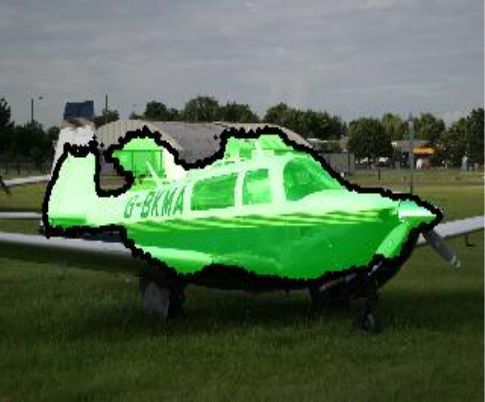}  & 
      \includegraphics[width=.07\linewidth]{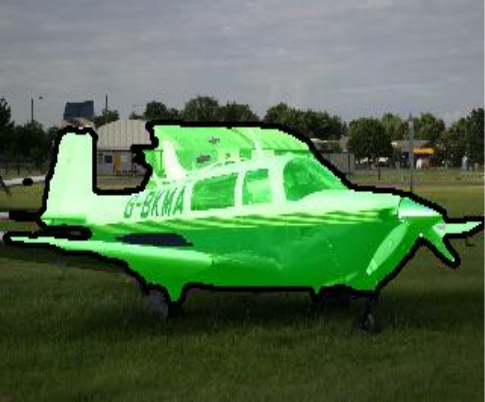}  & 
      \includegraphics[width=.07\linewidth]{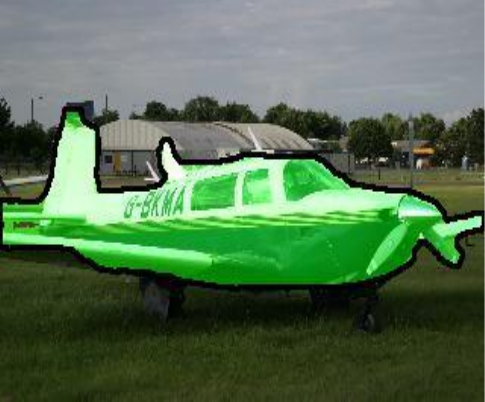} & 
      \multirow{2}*{Horse} &
      \includegraphics[width=.07\linewidth]{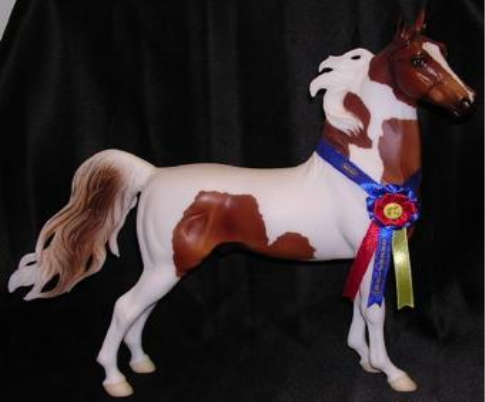} &
       \includegraphics[width=.07\linewidth]{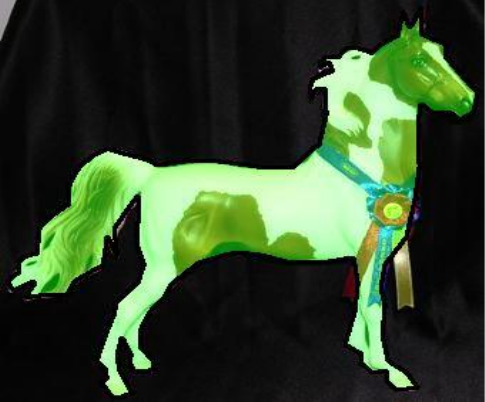} & \includegraphics[width=.07\linewidth]{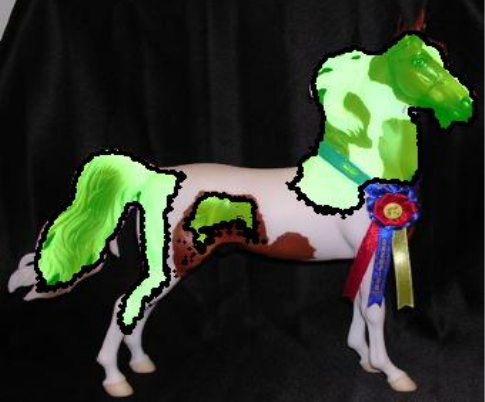} &
      \includegraphics[width=.07\linewidth]{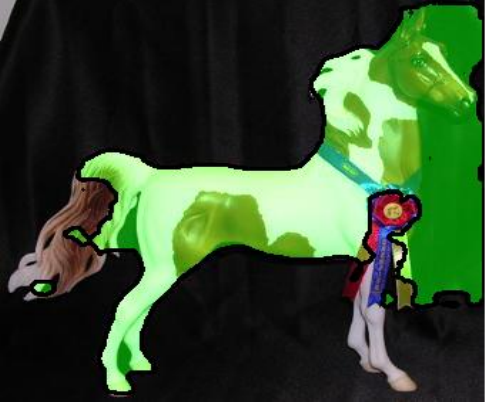} &
      \includegraphics[width=.07\linewidth]{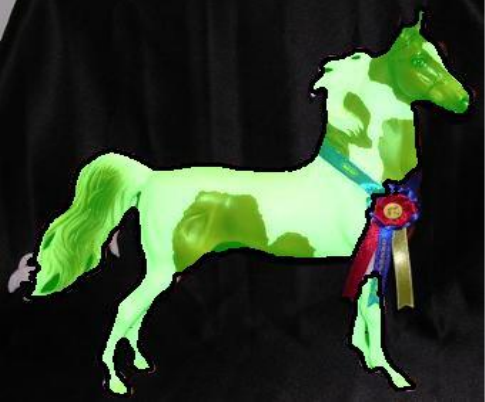} \\

       \multirow{2}*{} &
       \includegraphics[width=.07\linewidth]{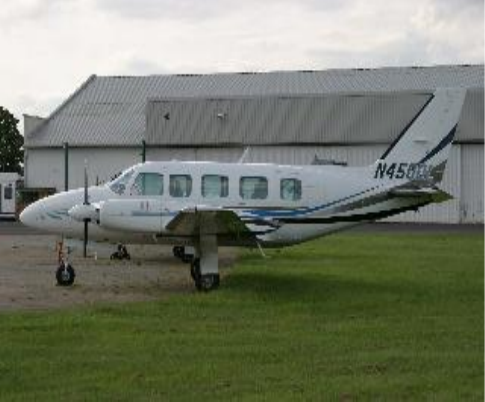}  & 
      \includegraphics[width=.07\linewidth]{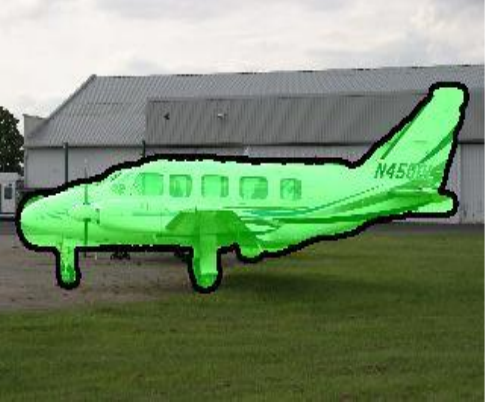}  & \includegraphics[width=.07\linewidth]{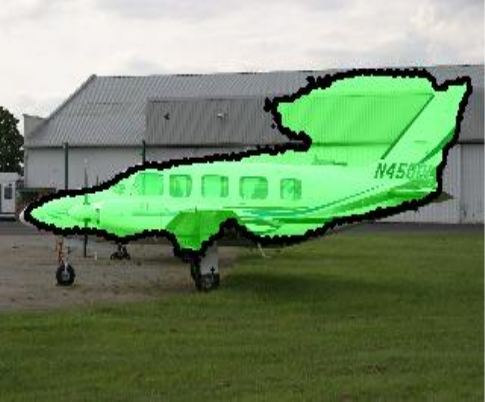}  & 
      \includegraphics[width=.07\linewidth]{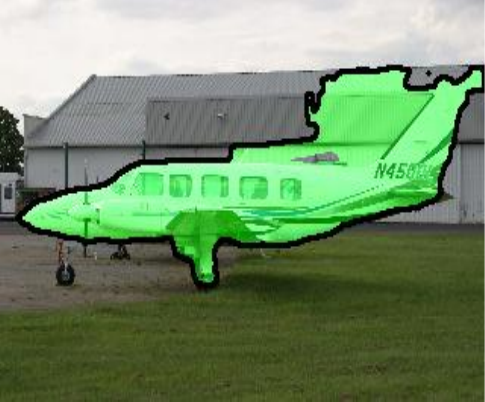}  & 
      \includegraphics[width=.07\linewidth]{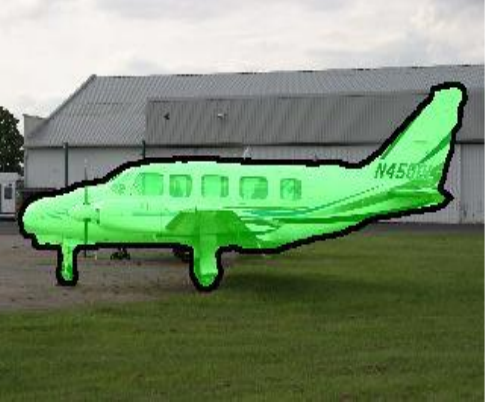} & 
       \multirow{2}*{} &
      \includegraphics[width=.07\linewidth]{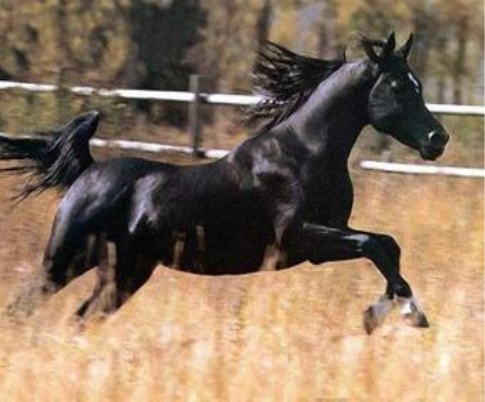} &
       \includegraphics[width=.07\linewidth]{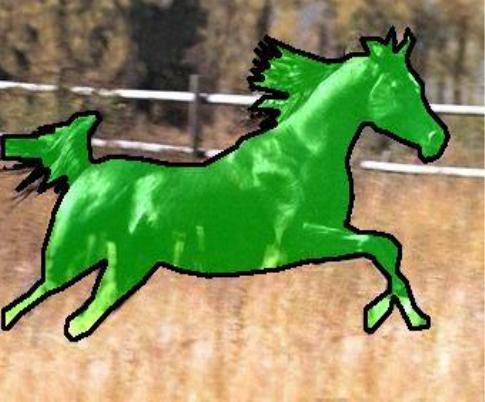} & \includegraphics[width=.07\linewidth]{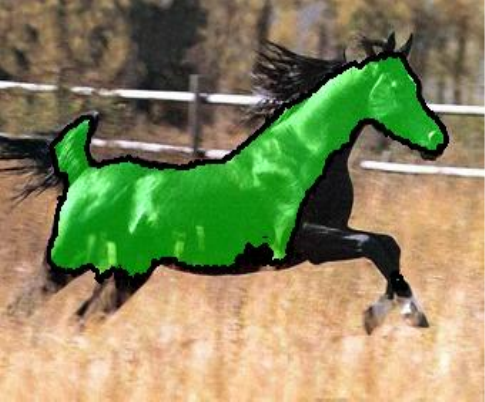} &
      \includegraphics[width=.07\linewidth]{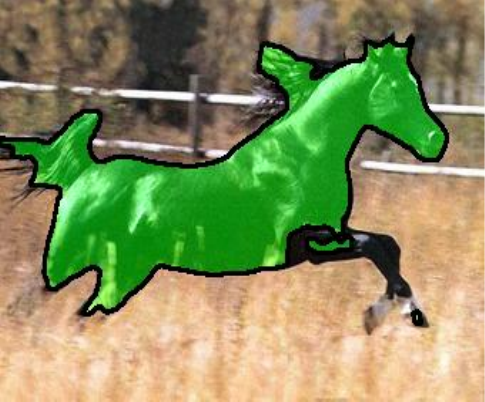} &
      \includegraphics[width=.07\linewidth]{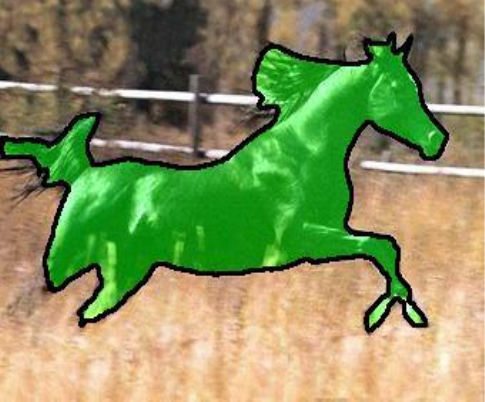} \\

        \multirow{2}*{Helicopter} &
       \includegraphics[width=.07\linewidth]{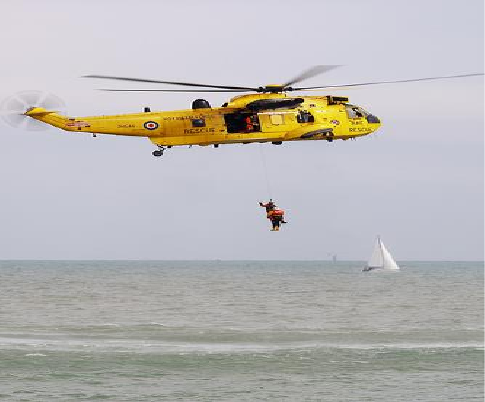}  & 
      \includegraphics[width=.07\linewidth]{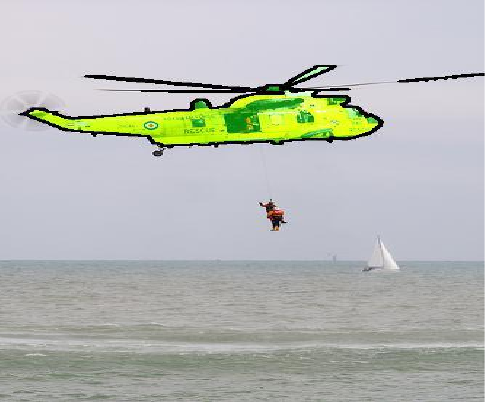}  & \includegraphics[width=.07\linewidth]{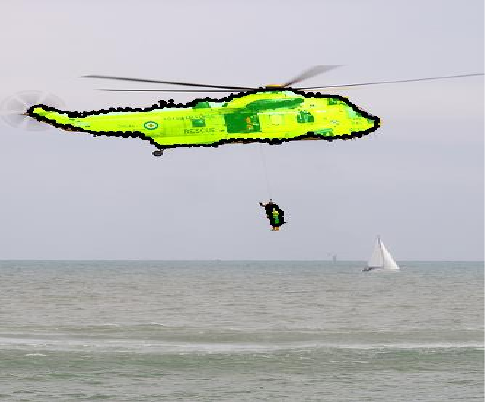}  & 
      \includegraphics[width=.07\linewidth]{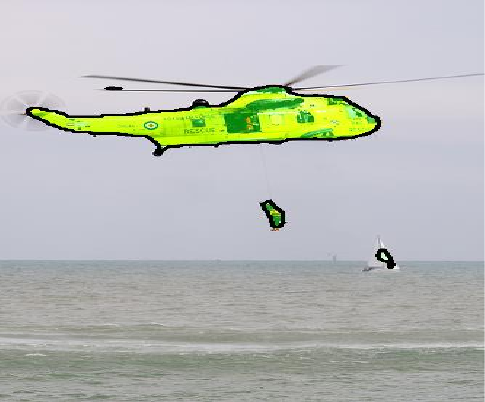}  & 
      \includegraphics[width=.07\linewidth]{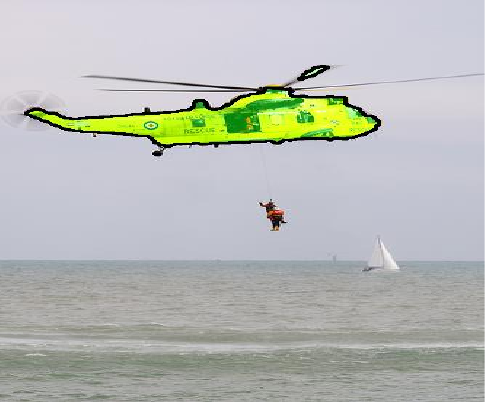} & 
      \multirow{2}*{Motorbike} &
      \includegraphics[width=.07\linewidth]{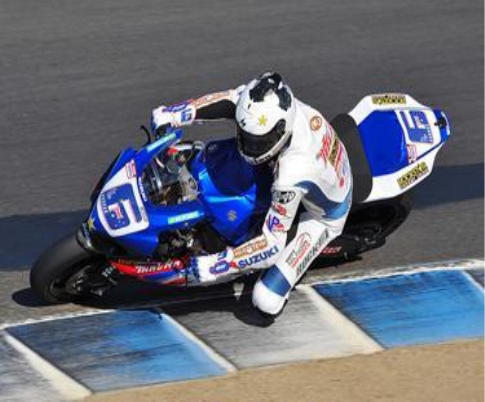} &
       \includegraphics[width=.07\linewidth]{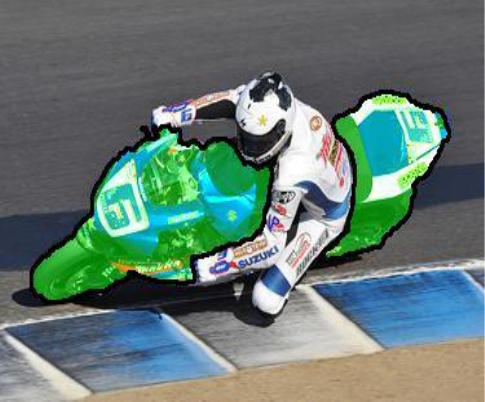} & \includegraphics[width=.07\linewidth]{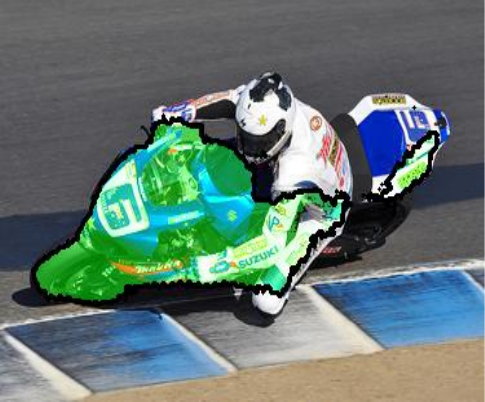} &
      \includegraphics[width=.07\linewidth]{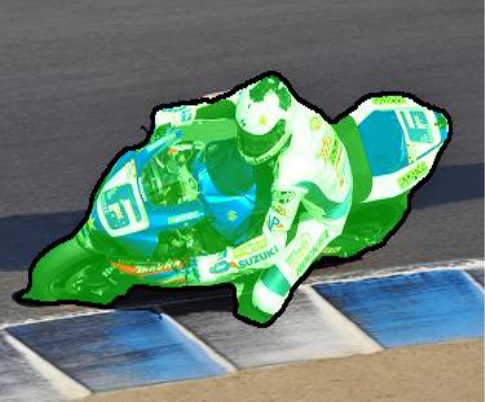} &
      \includegraphics[width=.07\linewidth]{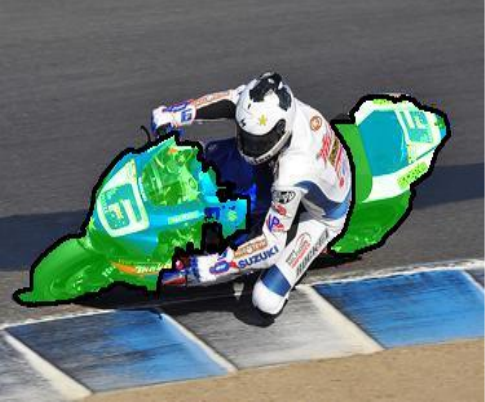} \\

      \multirow{2}*{} &
      \includegraphics[width=.07\linewidth]{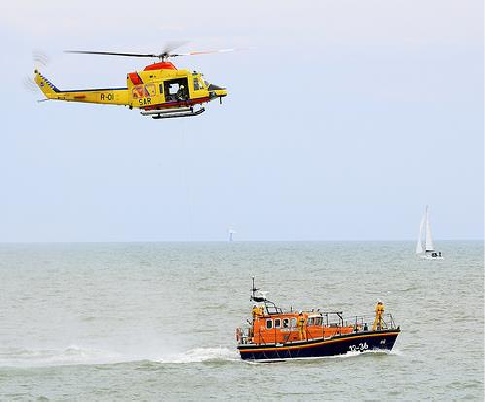}  & 
      \includegraphics[width=.07\linewidth]{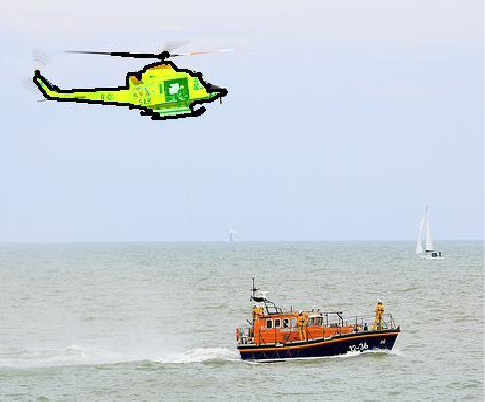}  & \includegraphics[width=.07\linewidth]{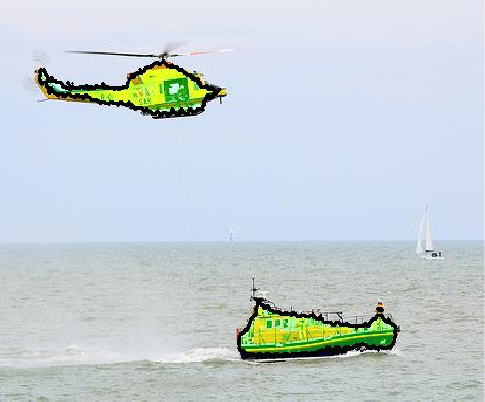}  & 
      \includegraphics[width=.07\linewidth]{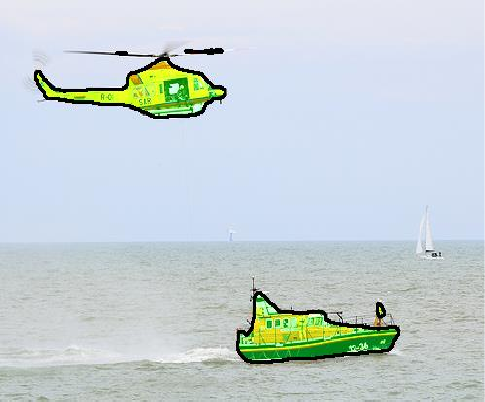}  & 
      \includegraphics[width=.07\linewidth]{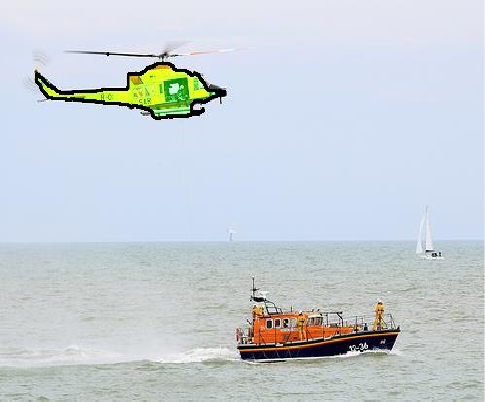} & 
      \multirow{2}*{} &
      \includegraphics[width=.07\linewidth]{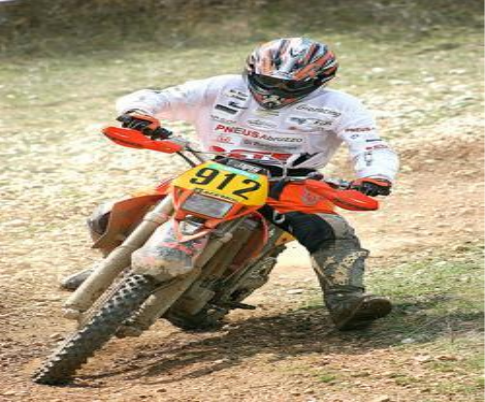} &
       \includegraphics[width=.07\linewidth]{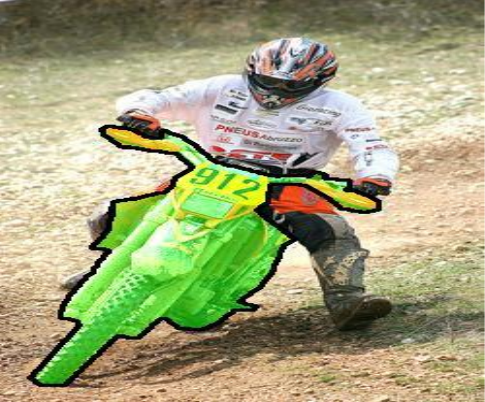} & \includegraphics[width=.07\linewidth]{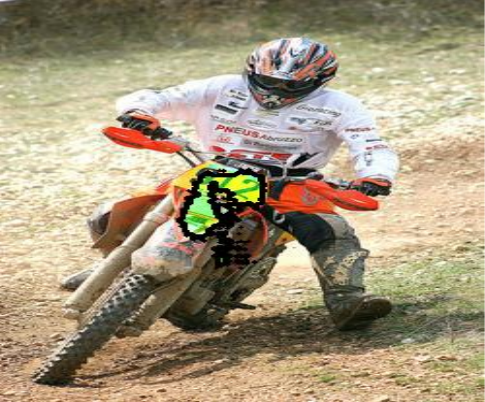} &
      \includegraphics[width=.07\linewidth]{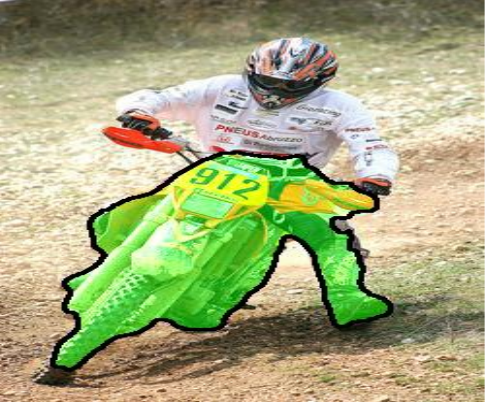} &
      \includegraphics[width=.07\linewidth]{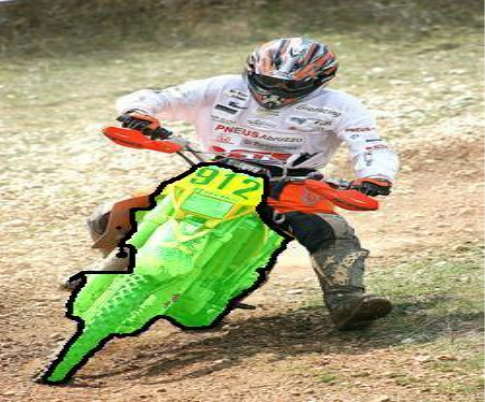} \\

        \multirow{2}*{Panda} &
       \includegraphics[width=.07\linewidth]{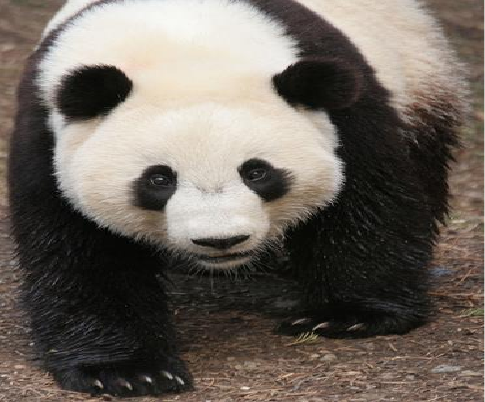}  & 
      \includegraphics[width=.07\linewidth]{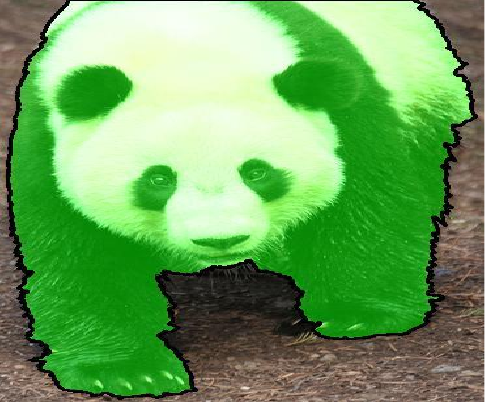}  & \includegraphics[width=.07\linewidth]{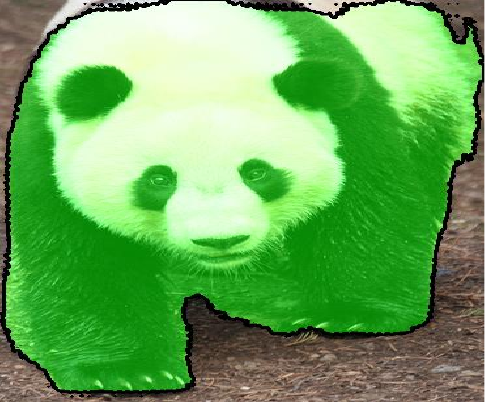}  & 
      \includegraphics[width=.07\linewidth]{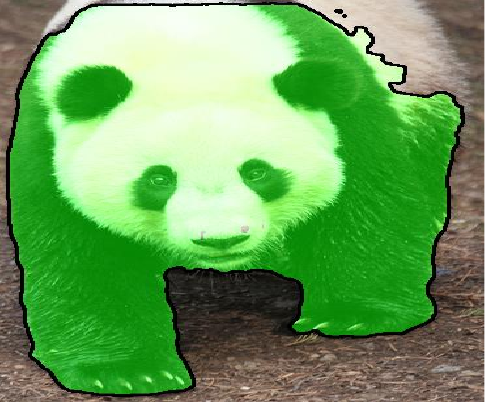}  & 
      \includegraphics[width=.07\linewidth]{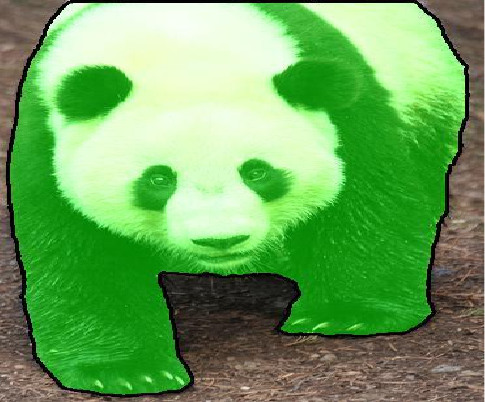} & 
       \multirow{2}*{Bus} &
      \includegraphics[width=.07\linewidth]{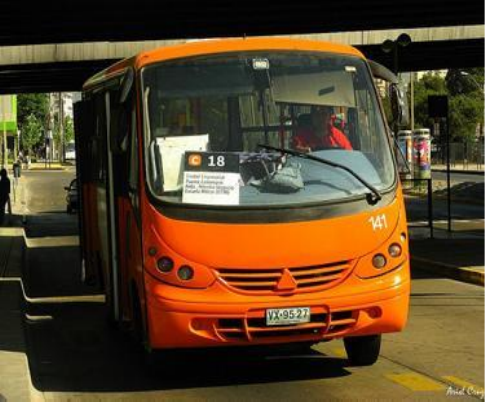} &
       \includegraphics[width=.07\linewidth]{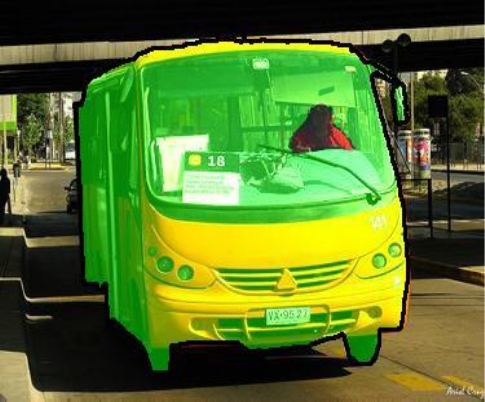} & \includegraphics[width=.07\linewidth]{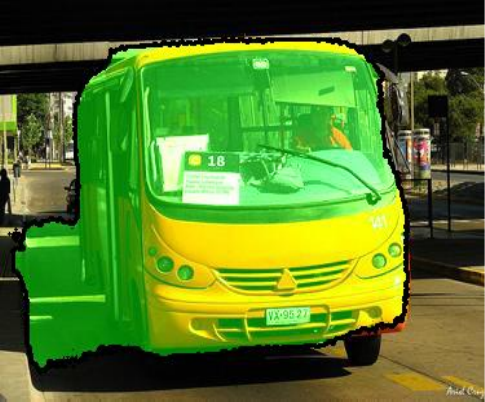} &
      \includegraphics[width=.07\linewidth]{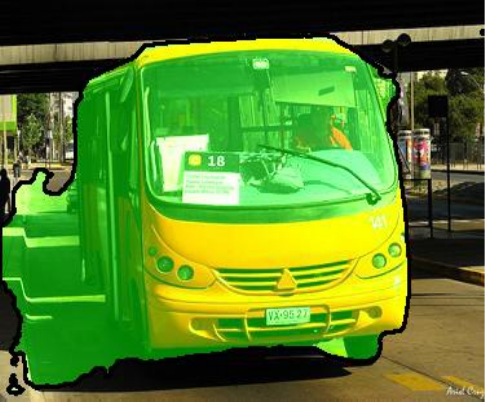} &
      \includegraphics[width=.07\linewidth]{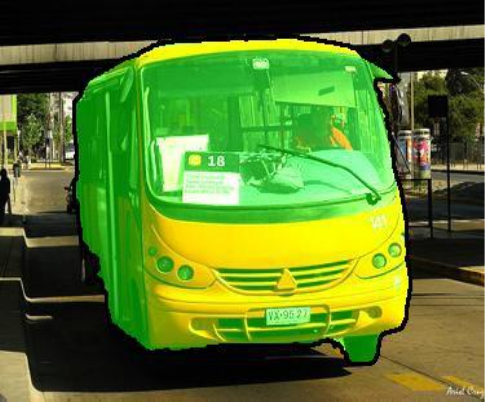} \\

        \multirow{2}*{} &
       \includegraphics[width=.07\linewidth]{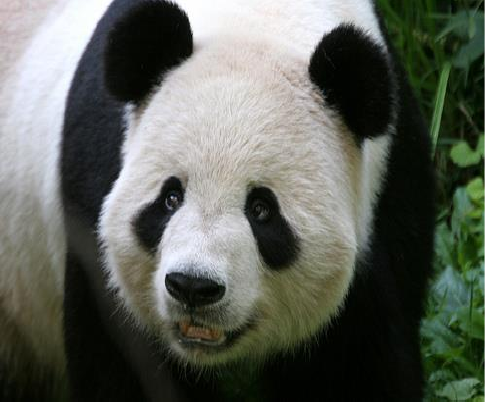}  & 
      \includegraphics[width=.07\linewidth]{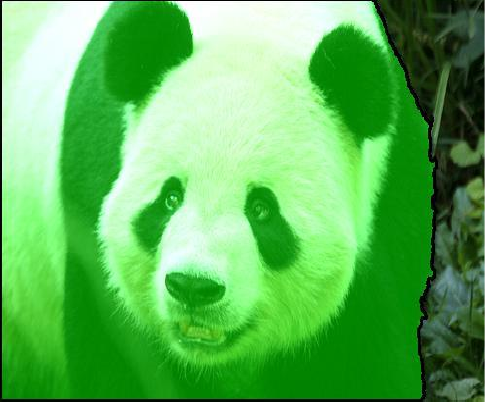}  & \includegraphics[width=.07\linewidth]{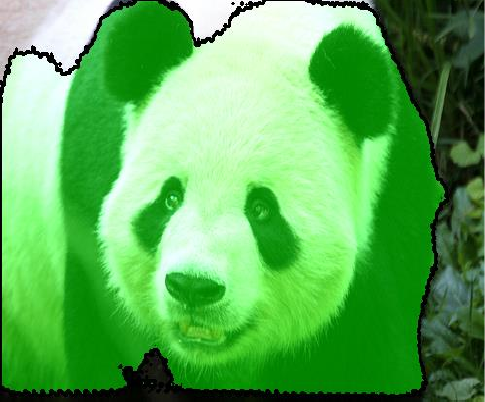}  & 
      \includegraphics[width=.07\linewidth]{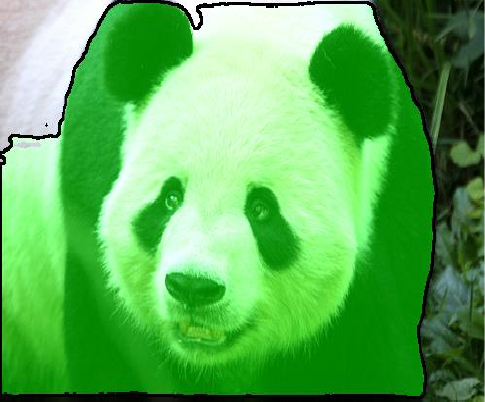}  & 
      \includegraphics[width=.07\linewidth]{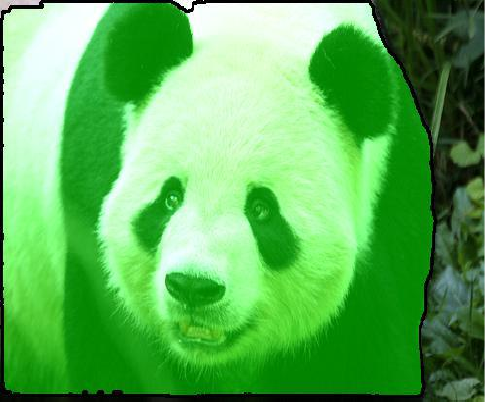} & 
      \multirow{2}*{} &
      \includegraphics[width=.07\linewidth]{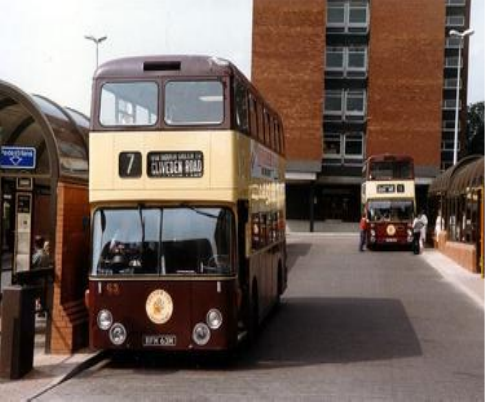} &
       \includegraphics[width=.07\linewidth]{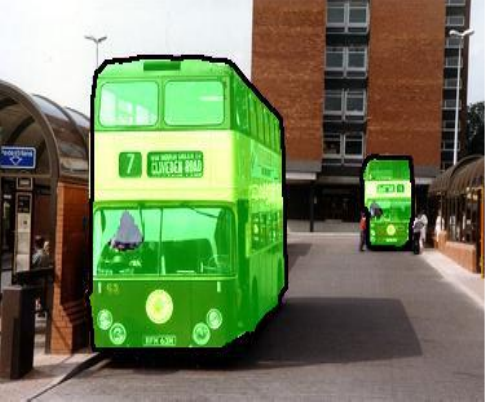} & \includegraphics[width=.07\linewidth]{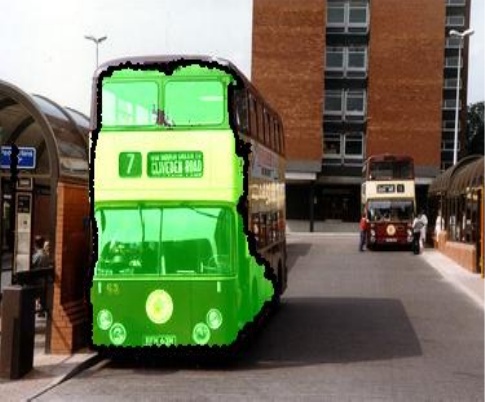} &
      \includegraphics[width=.07\linewidth]{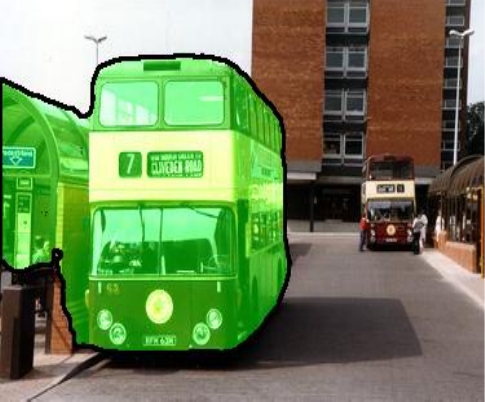} &
      \includegraphics[width=.07\linewidth]{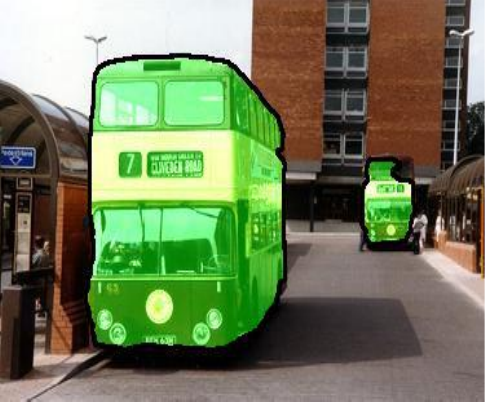} \\

        \small   &
        \small (a) Image 
        & \small (b) GT
        & \small (c) Zhang 
        & \small (d) Su  
        &  \small (e) Ours
        \small   &
        & \small (a) Image 
        & \small (b) GT 
        & \small (c) Zhang 
        & \small (d) Su  
        &  \small (e) Ours  \\
       
    \end{tabular}
    \end{subtable}
\end{center}

  \caption{\it \small Qualitative comparisons on the iCoseg (1$^{st}$-5$^{th}$ columns, 1$^{st}$-4$^{th}$ rows), Internet (6$^{th}$-10$^{th}$ columns, 1$^{st}$-4$^{th}$ rows) , MSRC (1$^{th}$-5$^{th}$ columns, 5$^{st}$-8$^{th}$ rows) and PASCAL (6$^{th}$-10$^{th}$ columns, 5$^{st}$-8$^{th}$ rows) datasets.  (a) Input images. (b) The ground-truth (GT) co-segmentation masks. (c) Predictions from Zhang \cite{zhang2020deep}. (d) Predictions from  Su \cite{su2022unified}. (e) Ours. 
  }
\label{segvis}
\vspace{-1.5em}
\end{figure*}

\paragraph{Classification Loss.} 
To optimize $\text{MLP}_3(\cdot)$ in Eq.~\eqref{eq:pscla}, we use ground-truth masks to compute the most likely semantic class within $P$ classes, and obtain the ground-truth one-hot similarity vector $\boldsymbol{\upsilon }^\text{gt}$. By minimizing the difference between estimated similarity vector $\boldsymbol{\upsilon }$ and $\boldsymbol{\upsilon }^\text{gt}$ using the Binary Cross-Entropy loss, we optimize $\text{MLP}_3(\cdot)$. The loss is given by,
\begin{equation}
    \mathcal{L}_{\text{c}} = - \frac{1}{P}\sum_{i=1}^{P} \upsilon_i^\text{gt} \log \hat{\upsilon_i} - (1 - \upsilon_i^\text{gt}) \log (1 - \hat{\upsilon_i}) \ .
\end{equation}
To compute the ground-truth most likely semantic class using $\{\mathbf{M}^\text{gt}_{i}\}_{i=1}^{N}$, we first segment images using their corresponding ground-truth masks, resulting in images of common semantic objects. Masked images are fed to the CLIP image encoder to obtain image embeddings $\mathcal{H}^{\tt{img}}_\text{gt}$, corresponding to the most common semantic. By computing pairwise cosine similarity between feature embeddings in  $\mathcal{H}^{\tt{img}}_\text{gt}$ and $\mathcal{H}^{\tt{txt}}$, we obtain a similarity matrix $\mathbf{S}^\text{gt} \in \mathbb{R}^{N \times P}$. By summarizing rows of $\mathbf{S}^\text{gt}$, we get the ground-truth similarity vector. The most likely semantic class is identified by finding the largest similarity within the vector.


\section{Experiments}

\paragraph{Datasets.} Following past methods \cite{zhang2021cyclesegnet}, we train our model on the training fold of PASCAL-VOC (PASCAL for short) \cite{pascal-voc-2012} or COCO \cite{lin2014microsoft} datasets, and test the trained model on MSRC \cite{shotton2006textonboost}, Internet \cite{rubinstein2013unsupervised}, and iCoseg \cite{batra2010icoseg}, and PASCAL (testing fold) datasets.

\paragraph{Evaluation Metrics.} We evaluate co-segmentation results of our model with Precision ($\mathcal{P}$) and Jaccard Index ($\mathcal{J}$) \cite{zhang2020deep}, the higher the better.

\paragraph{Implementations.} 
\label{sec:Implementations}
We use the ResNet50 \cite{he2016deep} as the backbone segmentation network and the pre-trained CLIP with a ViT-B/16 backbone. Please refer to the supplementary material for more implementation details.



\subsection{Comparison with State-of-the-arts}
The comparisons on the MSRC, Internet, iCoseg and PASCAL datasets are given in Tab.~\ref{icoseg}, respectively. Comparing with all methods, we achieve the best performance of Precision $\mathcal{P}$ and Jaccard Index $\mathcal{J}$ on the four datasets when training with COCO datasets. For methods trained on the PASCAL dataset, our method is comparable with respect to Zhang \cite{zhang2021cyclesegnet} on Precision $\mathcal{P}$ of the MSRC dataset, while outperforming the method on other evaluation metrics. For example, our methods have 1.2\% higher $\mathcal{J}$ than Zhang \cite{zhang2021cyclesegnet} on the MSRC dataset. In Fig.~\ref{segvis}, we qualitatively validate the effectiveness of our approach. The previous state-of-the-art methods fail to capture the accurate common semantics, resulting in sub-optimal co-segmentations in comparison to our methods.


\begin{table}[!t]
\centering
\caption{\it \small Ablation study of model components, ISFC (image set feature correspondence module), CLIP Inter. (CLIP interaction module), and CLIP Reg. (CLIP regularization module).
}
\label{tab:ablar}
\setlength{\tabcolsep}{7.5pt}
\begin{tabular}{ccccc}
\toprule
  ISFC  & CLIP Inter.  &  CLIP Reg.    & $\mathcal{P}$ (\%)  & $\mathcal{J}$ (\%)  \\ 
\midrule
 \multicolumn{3}{c}{Baseline}         & 95.7         & 86.6 \\ 
 \midrule
   \ding{52} &        &                & 96.2        & 88.3 \\ 
        &  \ding{52} &               & 96.6         & 90.0 \\ 
          &           &   \ding{52}    & 96.5        & 89.8 \\
   \ding{52} & \ding{52}  &            & 97.2        & 90.6 \\
   \ding{52}  &       &   \ding{52}    & 97.1        & 90.3 \\
          &  \ding{52} &    \ding{52}  & 97.4        & 90.8 \\
   \ding{52}  & \ding{52} &  \ding{52}  & \bf {97.8}   & \bf {91.7} \\
\bottomrule
\end{tabular}
\vspace{-1.5em}
\end{table}

\subsection{Ablation Studies}
\label{sec:abl}
Following \cite{zhang2021cyclesegnet}, in all ablation experiments, models are trained on the PASCAL dataset, and evaluated on the iCoseg dataset.

\paragraph{Ablation of Model Architectures.} The effectiveness of our model architectures is validated in Tab.~\ref{tab:ablar}, performing ablations on each proposed component, ISFC (image set feature correspondence module), CLIP Inter. (CLIP interaction module), and CLIP Reg. (CLIP regularization module). The `Baseline' setting independently feeds each image to the backbone network for segmentation, showing the lower bound performance of our co-segmentation task. We have the following findings:
i) each of our proposed components consistently improves the co-segmentation performance;
ii) with using all components, we have 2.1\% $\mathcal{P}$ and
5.1\% $\mathcal{J}$ improvements compared to using the `Baseline' setting.
We further qualitatively study our model components in Fig.~\ref{fig:moduls}. As shown, by gradually adding our ISFC, CLIP Inter., and CLIP Reg. modules to the `Baseline' setting, co-segmentation masks are refined in a coarse-to-fine manner.

\begin{figure}
\begin{center}
    \begin{subtable}{\linewidth}
    \setlength{\tabcolsep}{3pt}
    \begin{tabular}{cccccc}
      \includegraphics[width=.145\linewidth, height=.1\linewidth]{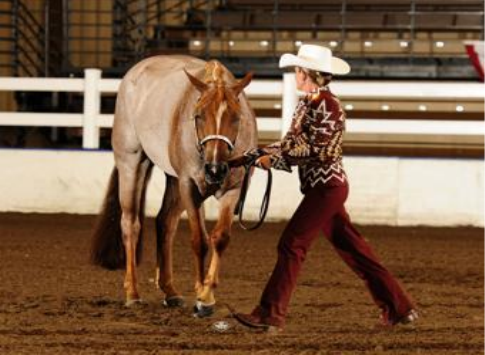}  &
      \includegraphics[width=.145\linewidth, height=.1\linewidth]{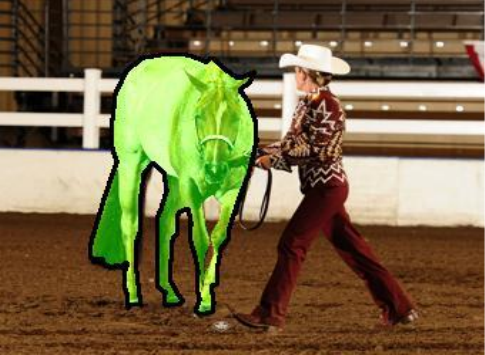}  &
      \includegraphics[width=.145\linewidth, height=.1\linewidth]{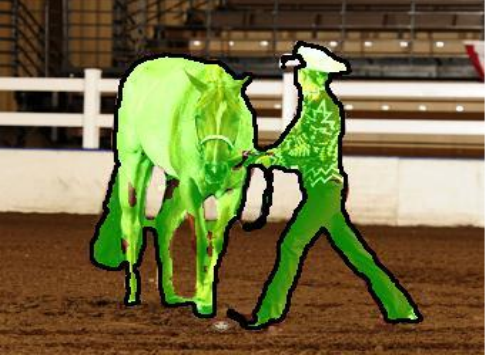}  &
      \includegraphics[width=.145\linewidth, height=.1\linewidth]{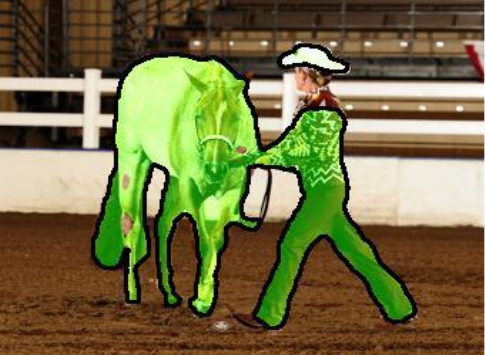}  &
      \includegraphics[width=.145\linewidth, height=.1\linewidth]{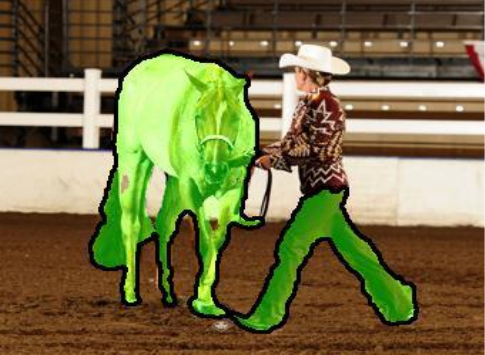}  &
       \includegraphics[width=.145\linewidth, height=.1\linewidth]{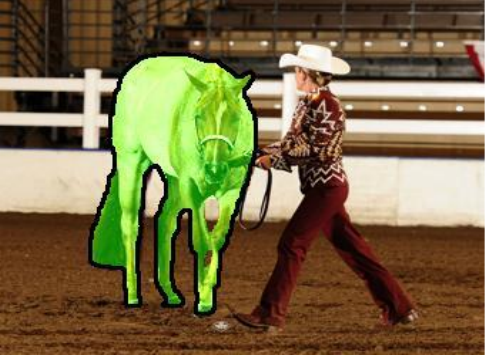}  \\

      \includegraphics[width=.145\linewidth, height=.1\linewidth]{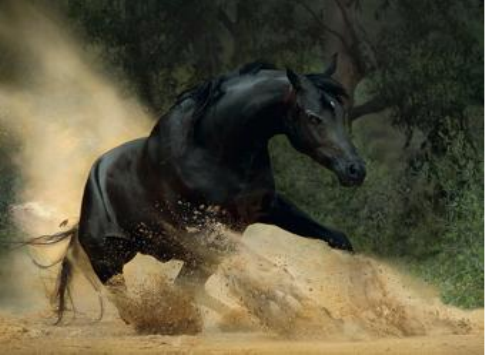}  &
      \includegraphics[width=.145\linewidth, height=.1\linewidth]{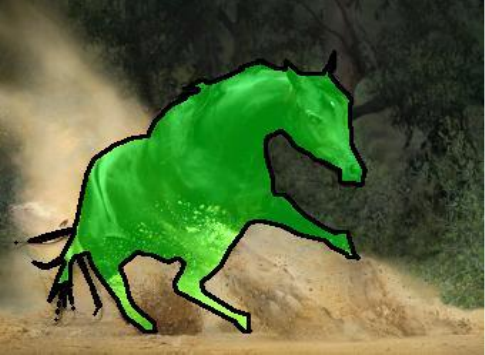}  &
      \includegraphics[width=.145\linewidth, height=.1\linewidth]{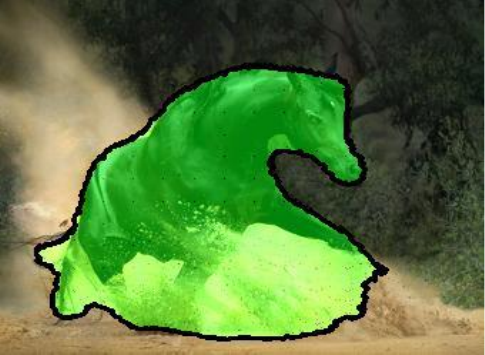}  &
      \includegraphics[width=.145\linewidth, height=.1\linewidth]{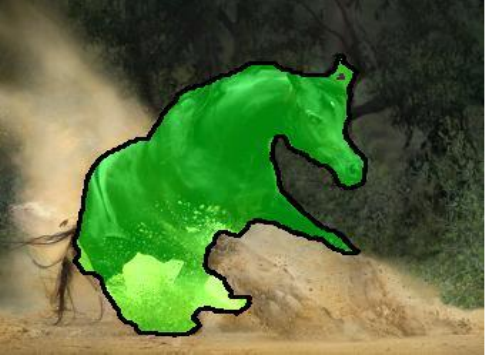}  &
      \includegraphics[width=.145\linewidth, height=.1\linewidth]{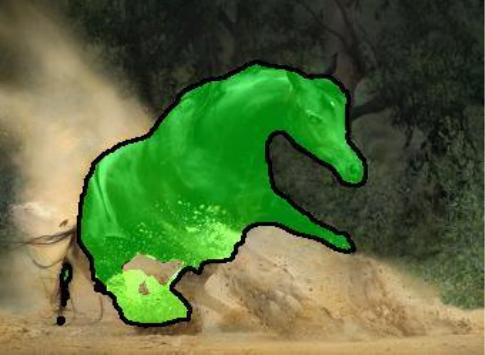}  &
       \includegraphics[width=.145\linewidth, height=.1\linewidth]{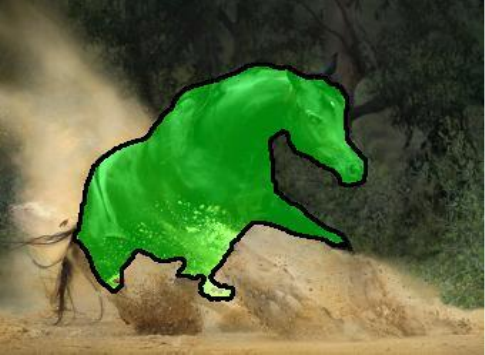}  \\

      \includegraphics[width=.145\linewidth, height=.1\linewidth]{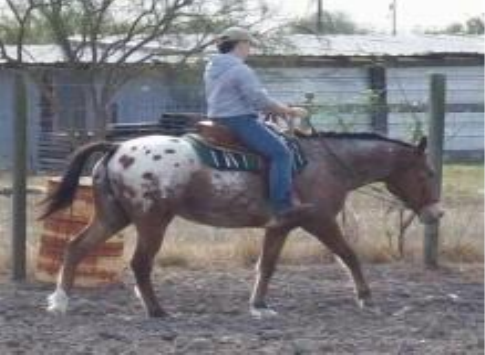}  &
      \includegraphics[width=.145\linewidth, height=.1\linewidth]{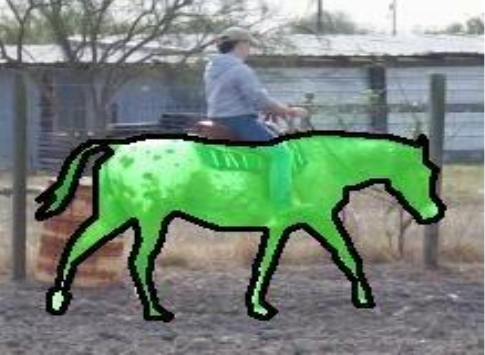}  &
      \includegraphics[width=.145\linewidth, height=.1\linewidth]{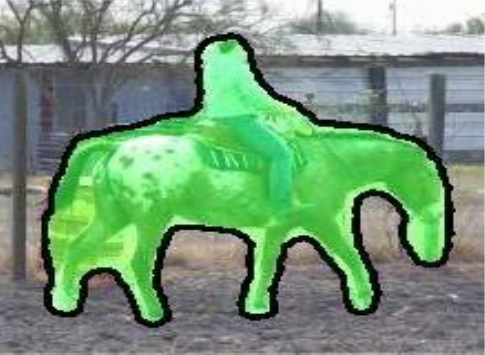}  &
      \includegraphics[width=.145\linewidth, height=.1\linewidth]{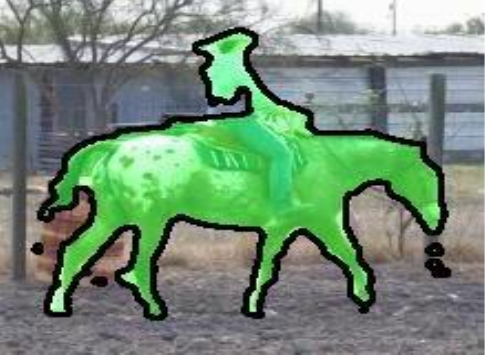}  &
      \includegraphics[width=.145\linewidth, height=.1\linewidth]{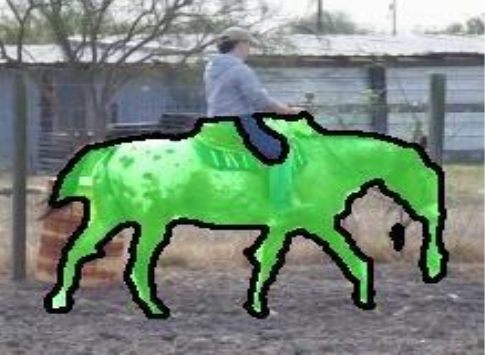}  &
       \includegraphics[width=.145\linewidth, height=.1\linewidth]{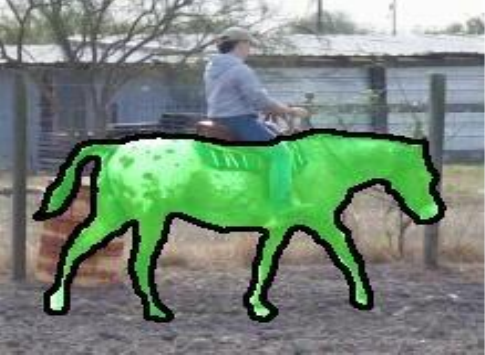}  \\

       \includegraphics[width=.145\linewidth, height=.1\linewidth]{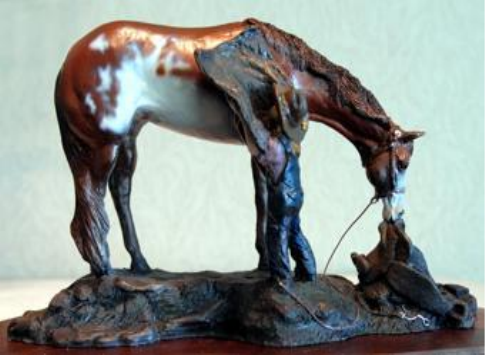}  &
      \includegraphics[width=.145\linewidth, height=.1\linewidth]{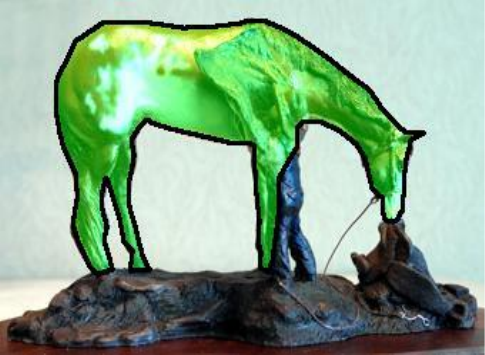}  &
      \includegraphics[width=.145\linewidth, height=.1\linewidth]{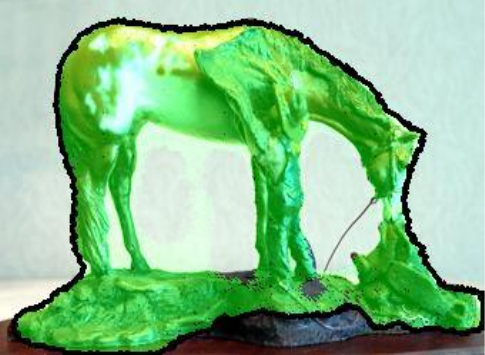}  &
      \includegraphics[width=.145\linewidth, height=.1\linewidth]{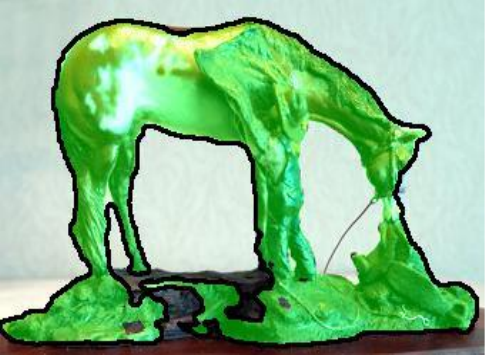}  &
      \includegraphics[width=.145\linewidth, height=.1\linewidth]{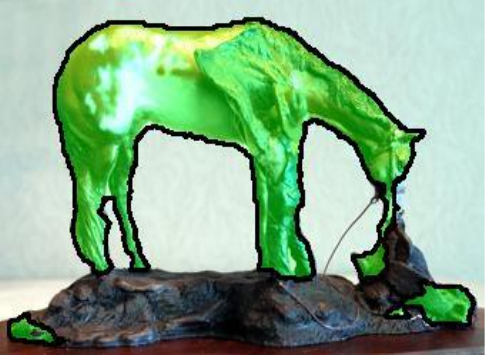}  &
       \includegraphics[width=.145\linewidth, height=.1\linewidth]{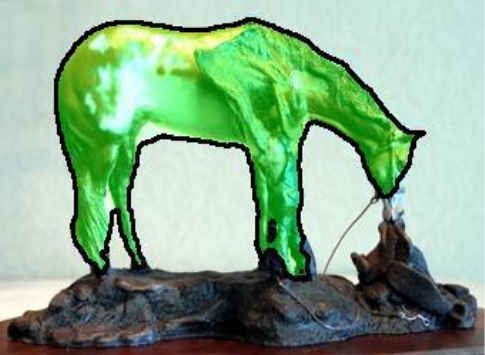}  \\

       \includegraphics[width=.145\linewidth, height=.1\linewidth]{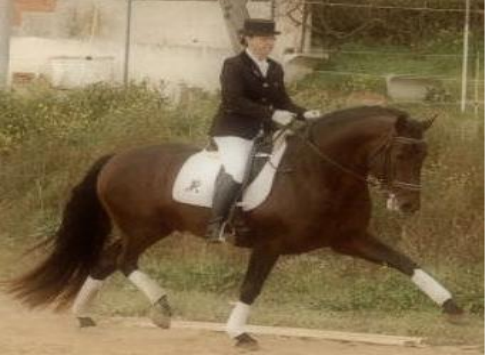}  &
      \includegraphics[width=.145\linewidth, height=.1\linewidth]{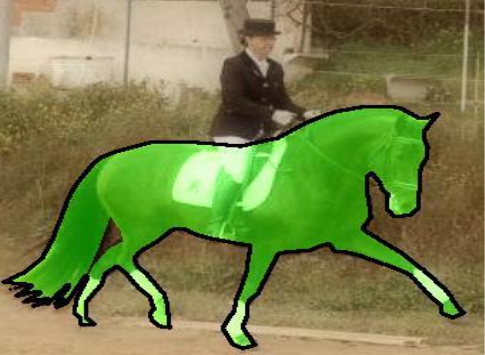}  &
      \includegraphics[width=.145\linewidth, height=.1\linewidth]{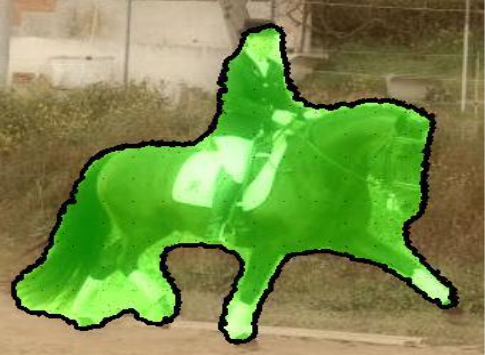}  &
      \includegraphics[width=.145\linewidth, height=.1\linewidth]{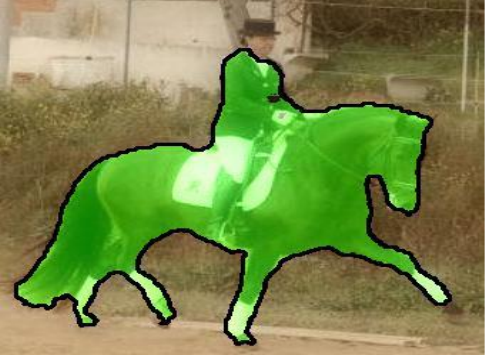}  &
      \includegraphics[width=.145\linewidth, height=.1\linewidth]{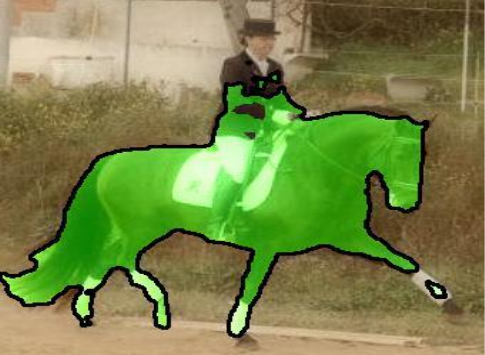}  &
       \includegraphics[width=.145\linewidth, height=.1\linewidth]{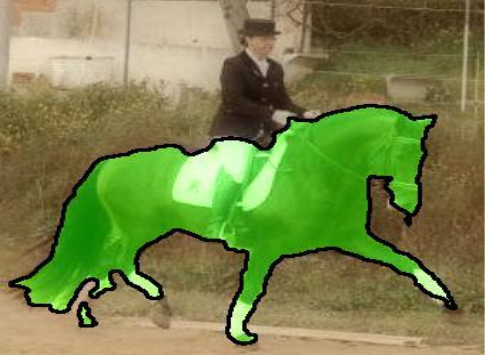}  \\

        \small (a)  
        & \small (b)  
        & \small (c)   
        &  \small (d) 
        & \small (e) 
        & \small (f) \\

 \end{tabular}

    \end{subtable}
\end{center}
  \caption{\it \small \it Qualitative comparisons of proposed modules. (a) Input images. (b) Ground-truth masks. (c) Predictions from baseline. (d) Predictions with the ISFC module. (e) Predictions with ISFC and CLIP Inter. modules. (f) Ours that with three key modules. }
\label{fig:moduls}
\vspace{-.5em}
\end{figure}

\begin{table}[!t]
\centering
\caption{\it \small Analysis of losses.
}

\setlength{\tabcolsep}{14pt}
\begin{tabular}{ccccc}
\toprule
$\mathcal{L}_{\text{iou}}$ &  $\mathcal{L}_{\text{cs}}$  &  $\mathcal{L}_{\text{c}}$  & $\mathcal{P}$ (\%) & $\mathcal{J}$ (\%) \\  
\midrule
\ding{52}  &                &            &    97.4       &   90.9        \\
\ding{52}  &  \ding{52}     &            &    97.6       &   91.2        \\
\ding{52}  &                &  \ding{52} &    97.6       &   91.3        \\
\ding{52}  &  \ding{52}     &  \ding{52} &    \textbf{97.8}      &   \textbf{91.7}        \\ 
\bottomrule
\end{tabular}
\label{tab:los}
\end{table}

\paragraph{Ablation of Losses.} We study the optimization losses in Tab.~\ref{tab:los}. With using the $\mathcal{L}_{\text{iou}}$, we have 97.4\% $\mathcal{P}$ and 90.9\% $\mathcal{J}$. By using $\mathcal{L}_{\text{cs}}$ or $\mathcal{L}_{\text{c}}$, there are 0.2\%/0.3\% and 0.2\%/0.4\% higher $\mathcal{P}$ and $\mathcal{J}$.  Combining all loses, we have the best performance, achieving 97.8\% $\mathcal{P}$ and 91.7\% $\mathcal{J}$.

\paragraph{Distilled CLIP Text Embeddings.} We study the impact of the number (${K}$) of distilled CLIP text embeddings in our CLIP interaction module (Tab.~\ref{tab:top}), \ie, $\text{TopK}(\boldsymbol{\sigma})$. We find that the performance of our method saturates when we have ${K}\geq5$.
We therefore set ${K}=5$ for all experiments. 

\begin{table}[!t]
\centering
\caption{\it \small Analysis of the number (${K}$) for distilled text embeddings.}
\setlength{\tabcolsep}{29pt}
\begin{tabular}{ccc}
   \toprule
    $K$ & $\mathcal{P}$ (\%) & $\mathcal{J}$ (\%) \\  
    \midrule
     1  &   97.72        &   91.53        \\
    3  &   97.79        &   91.67        \\
    5  &   97.83        &  91.74         \\
    7   &  97.83         &  91.73         \\
    9  &   97.82        &    91.74       \\
    \bottomrule
\end{tabular}
\label{tab:top}
\end{table}

\subsection{Discussions}

\paragraph{More Images.} 

Testing the model with a different number of co-segmentation images used in training potentially causes domain shifts and noise to model inference. For example, the runner-up method, Su \cite{su2022unified}, loses 1.3 \% $\mathcal{P}$ and 2.4\% $\mathcal{J}$ when increasing the number of input images from 5 to 8.

However, in our model, we consistently improve performance if use more images, having
0.3\% $\mathcal{P}$ and 0.2\% $\mathcal{J}$ improvements 
by using 8 images.

\paragraph{Comparison with Foundation Models.}
\label{sec:Large_Segmentation_Models} 
We compare with foundation segmentation models in the co-segmentation tasks. There are four settings explored:
i) `SAM'. We use the automatic segmentation mask generators from SAM \cite{kirillov2023segany}. With the masks, we segment the images and leverage the zero-shot classification ability of CLIP to find the common semantics \cite{xie2023edit};
ii) `SAM GT'. We provide the bounding boxes calculated from the ground-truth co-segmentation masks to SAM, studying the upper bound performance of SAM in our task;
iii) `SEEM'. We use SEEM \cite{zou2023segment} to automatically segment the images, and classify them into different classes. The class with majority votes is used to choose the co-segmentation masks;
iv) `SEEM GT'. The ground-truth common semantics are provided for choosing the masks predicted by SEEM; 
v) `Ours GT'. We provide the ground-truth common semantics to our model. Even compared with these large foundation models that are supplied with ground-truth information, our model has the best results. Meanwhile, our model has almost the same performance as `Ours GT', validating our assumption of completeness on $\mathcal{H}^{\tt{txt}}$ and the effectiveness of our soft text semantic distillation module, though $\mathcal{H}^{\tt{txt}}$ is not the same as the dataset common semantics. 

\begin{table}[!t]
\begin{center}
\caption{\it \small Comparison with foundation segmentation models. }

\setlength{\tabcolsep}{23pt}
\begin{tabular}{lcc}
\toprule
Method &           $\mathcal{P}$ (\%) &   $\mathcal{J}$ (\%) \\
\midrule
SAM &   69.17  &  35.30   \\
SAM GT &   97.35  &  92.01  \\
SEEM &  95.78  &  74.26   \\
SEEM GT &  97.25  &  88.32   \\
\midrule
\textbf{Ours}   &   \textbf{98.31} & \textbf{92.92}\\
\textbf{Ours GT}   &   \textbf{98.34} & \textbf{92.93}\\
\bottomrule
\end{tabular}
\end{center}
\label{LSM}
\end{table}

\paragraph{Limitations.} Compared to past methods, our framework uses large-scale CLIP models. While leveraging the strong semantic discovery ability of CLIP, extra computes are scarified. However, our method is still computationally efficient. For example, when comparing with the most competitive past method, our method is faster than Zhang \cite{zhang2020deep} and Su \cite{su2022unified} on an NVIDIA 3090 GPU, even the two methods use ground-truth common semantics in training which potentially leads to more lightweight network weights.

\section{Conclusions}
We propose a new method for the image co-segmentation task by leveraging the powerful zero-shot ability of CLIP to extract semantic information. 
We propose i) an image set feature correspondence module, encoding global consistent semantic information of the image set; ii) a CLIP interaction module, modulating the intermediate backbone segmentation features with Top-K common  CLIP semantics; iii) a CLIP regularization module, identifying the most common semantic object for the image set. We use the most common semantic to regularize backbone segmentation features. Our network is trained end-to-end, with two new proposed segmentation and classification losses. Experiments on four standard image co-segmentation benchmark datasets demonstrate the state-of-the-art performance of our method.

\paragraph{Acknowledgment.} The work was supported in part by the Beijing Institute of Technology Research Fund Program for Young Scholars

\bibliography{aaai24}

\begin{thebibliography}{40}
\providecommand{\natexlab}[1]{#1}

\bibitem[{Banerjee et~al.(2019)Banerjee, Hati, Chaudhuri, and
  Velmurugan}]{banerjee2019cosegnet}
Banerjee, S.; Hati, A.; Chaudhuri, S.; and Velmurugan, R. 2019.
\newblock CoSegNet: Image Co-segmentation using a Conditional Siamese
  Convolutional Network.
\newblock In \emph{IJCAI}, 673--679.

\bibitem[{Batra et~al.(2010)Batra, Kowdle, Parikh, Luo, Chen
  et~al.}]{batra2010icoseg}
Batra, D.; Kowdle, A.; Parikh, D.; Luo, J.; Chen, T.; et~al. 2010.
\newblock icoseg: Interactive co-segmentation with intelligent scribble
  guidance.
\newblock In \emph{2010 IEEE computer society conference on computer vision and
  pattern recognition}, 3169--3176. IEEE.

\bibitem[{Chen et~al.(2018)Chen, Huang, Nakayama et~al.}]{chen2018semantic}
Chen, H.; Huang, Y.; Nakayama, H.; et~al. 2018.
\newblock Semantic aware attention based deep object co-segmentation.
\newblock In \emph{Asian Conference on Computer Vision}, 435--450. Springer.

\bibitem[{Ding et~al.(2022)Ding, Xue, Xia, Dai et~al.}]{ding2022decoupling}
Ding, J.; Xue, N.; Xia, G.-S.; Dai, D.; et~al. 2022.
\newblock Decoupling zero-shot semantic segmentation.
\newblock In \emph{Proceedings of the IEEE/CVF Conference on Computer Vision
  and Pattern Recognition}, 11583--11592.

\bibitem[{Everingham et~al.(2012)Everingham, Van~Gool, Williams, Winn, and
  Zisserman}]{pascal-voc-2012}
Everingham, M.; Van~Gool, L.; Williams, C. K.~I.; Winn, J.; and Zisserman, A.
  2012.
\newblock The {PASCAL} {V}isual {O}bject {C}lasses {C}hallenge 2012 {(VOC2012)}
  {R}esults.
\newblock
  ``http://www.pascal-network.org/challenges/VOC/voc2012/workshop/index.html".

\bibitem[{Faktor, Irani et~al.(2013)}]{faktor2013co}
Faktor, A.; Irani, M.; et~al. 2013.
\newblock Co-segmentation by composition.
\newblock In \emph{Proceedings of the IEEE international conference on computer
  vision}, 1297--1304.

\bibitem[{Hamilton et~al.(2017)Hamilton, Ying, Leskovec
  et~al.}]{hamilton2017inductive}
Hamilton, W.; Ying, Z.; Leskovec, J.; et~al. 2017.
\newblock Inductive representation learning on large graphs.
\newblock \emph{Advances in neural information processing systems}, 30.

\bibitem[{Han et~al.(2017)Han, Quan, Zhang, and Nie}]{han2017robust}
Han, J.; Quan, R.; Zhang, D.; and Nie, F. 2017.
\newblock Robust object co-segmentation using background prior.
\newblock \emph{IEEE Transactions on Image Processing}, 27(4): 1639--1651.

\bibitem[{He et~al.(2016)He, Zhang, Ren, Sun et~al.}]{he2016deep}
He, K.; Zhang, X.; Ren, S.; Sun, J.; et~al. 2016.
\newblock Deep residual learning for image recognition.
\newblock In \emph{Proceedings of the IEEE conference on computer vision and
  pattern recognition}, 770--778.

\bibitem[{Hsu et~al.(2018)Hsu, Lin, Chuang et~al.}]{hsu2018co}
Hsu, K.-J.; Lin, Y.-Y.; Chuang, Y.-Y.; et~al. 2018.
\newblock Co-attention CNNs for unsupervised object co-segmentation.
\newblock In \emph{IJCAI}, volume~1, 2.

\bibitem[{Jerripothula et~al.(2016)Jerripothula, Cai, Yuan
  et~al.}]{jerripothula2016image}
Jerripothula, K.~R.; Cai, J.; Yuan, J.; et~al. 2016.
\newblock Image co-segmentation via saliency co-fusion.
\newblock volume~18, 1896--1909. IEEE.

\bibitem[{Kirillov et~al.(2023)Kirillov, Mintun, Ravi, Mao, Rolland, Gustafson,
  Xiao, Whitehead, Berg, Lo, Doll{\'a}r, Girshick et~al.}]{kirillov2023segany}
Kirillov, A.; Mintun, E.; Ravi, N.; Mao, H.; Rolland, C.; Gustafson, L.; Xiao,
  T.; Whitehead, S.; Berg, A.~C.; Lo, W.-Y.; Doll{\'a}r, P.; Girshick, R.;
  et~al. 2023.
\newblock Segment Anything.
\newblock \emph{arXiv:2304.02643}.

\bibitem[{Li et~al.(2019)Li, Sun, Li, Wu, Hu et~al.}]{li2019group}
Li, B.; Sun, Z.; Li, Q.; Wu, Y.; Hu, A.; et~al. 2019.
\newblock Group-wise deep object co-segmentation with co-attention recurrent
  neural network.
\newblock In \emph{Proceedings of the IEEE/CVF International Conference on
  Computer Vision}, 8519--8528.

\bibitem[{Li et~al.(2018)Li, Hosseini~Jafari, Rother et~al.}]{li2018deep}
Li, W.; Hosseini~Jafari, O.; Rother, C.; et~al. 2018.
\newblock Deep object co-segmentation.
\newblock In \emph{Computer Vision--ACCV 2018: 14th Asian Conference on
  Computer Vision, Perth, Australia, December 2--6, 2018, Revised Selected
  Papers, Part III 14}, 638--653. Springer.

\bibitem[{Liang et~al.(2017)Liang, Zhu, Huang et~al.}]{liang2017multi}
Liang, X.; Zhu, L.; Huang, D.-S.; et~al. 2017.
\newblock Multi-task ranking SVM for image cosegmentation.
\newblock \emph{Neurocomputing}, 247: 126--136.

\bibitem[{Lin et~al.(2014)Lin, Maire, Belongie, Hays, Perona, Ramanan,
  Doll{\'a}r, Zitnick et~al.}]{lin2014microsoft}
Lin, T.-Y.; Maire, M.; Belongie, S.; Hays, J.; Perona, P.; Ramanan, D.;
  Doll{\'a}r, P.; Zitnick, C.~L.; et~al. 2014.
\newblock Microsoft coco: Common objects in context.
\newblock In \emph{Computer Vision--ECCV 2014: 13th European Conference,
  Zurich, Switzerland, September 6-12, 2014, Proceedings, Part V 13}, 740--755.
  Springer.

\bibitem[{Liu et~al.(2020)Liu, Zhang, Lin, Hung, Miao et~al.}]{liu2020weakly}
Liu, W.; Zhang, C.; Lin, G.; Hung, T.-Y.; Miao, C.; et~al. 2020.
\newblock Weakly supervised segmentation with maximum bipartite graph matching.
\newblock In \emph{Proceedings of the 28th ACM International Conference on
  Multimedia}, 2085--2094.

\bibitem[{Mustafa, Hilton et~al.(2017)}]{mustafa2017semantically}
Mustafa, A.; Hilton, A.; et~al. 2017.
\newblock Semantically coherent co-segmentation and reconstruction of dynamic
  scenes.
\newblock In \emph{Proceedings of the IEEE Conference on Computer Vision and
  Pattern Recognition}, 422--431.

\bibitem[{Quan et~al.(2016)Quan, Han, Zhang, Nie et~al.}]{quan2016object}
Quan, R.; Han, J.; Zhang, D.; Nie, F.; et~al. 2016.
\newblock Object co-segmentation via graph optimized-flexible manifold ranking.
\newblock In \emph{Proceedings of the IEEE conference on computer vision and
  pattern recognition}, 687--695.

\bibitem[{Radford et~al.(2021)Radford, Kim, Hallacy, Ramesh, Goh, Agarwal,
  Sastry, Askell, Mishkin, Clark et~al.}]{radford2021learning}
Radford, A.; Kim, J.~W.; Hallacy, C.; Ramesh, A.; Goh, G.; Agarwal, S.; Sastry,
  G.; Askell, A.; Mishkin, P.; Clark, J.; et~al. 2021.
\newblock Learning transferable visual models from natural language
  supervision.
\newblock In \emph{International conference on machine learning}, 8748--8763.
  PMLR.

\bibitem[{Rubinstein et~al.(2013)Rubinstein, Joulin, Kopf, Liu
  et~al.}]{rubinstein2013unsupervised}
Rubinstein, M.; Joulin, A.; Kopf, J.; Liu, C.; et~al. 2013.
\newblock Unsupervised joint object discovery and segmentation in internet
  images.
\newblock In \emph{Proceedings of the IEEE conference on computer vision and
  pattern recognition}, 1939--1946.

\bibitem[{Sarlin et~al.(2020)Sarlin, DeTone, Malisiewicz, Rabinovich
  et~al.}]{sarlin2020superglue}
Sarlin, P.-E.; DeTone, D.; Malisiewicz, T.; Rabinovich, A.; et~al. 2020.
\newblock Superglue: Learning feature matching with graph neural networks.
\newblock In \emph{Proceedings of the IEEE/CVF conference on computer vision
  and pattern recognition}, 4938--4947.

\bibitem[{Shen et~al.(2022)Shen, Efros, Joulin, Aubry
  et~al.}]{shen2022learning}
Shen, X.; Efros, A.~A.; Joulin, A.; Aubry, M.; et~al. 2022.
\newblock Learning co-segmentation by segment swapping for retrieval and
  discovery.
\newblock In \emph{Proceedings of the IEEE/CVF Conference on Computer Vision
  and Pattern Recognition}, 5082--5092.

\bibitem[{Shotton et~al.(2006)Shotton, Winn, Rother, Criminisi
  et~al.}]{shotton2006textonboost}
Shotton, J.; Winn, J.; Rother, C.; Criminisi, A.; et~al. 2006.
\newblock Textonboost: Joint appearance, shape and context modeling for
  mulit-class object recognition and segmentation.
\newblock In \emph{European conference on computer vision (ECCV)}.

\bibitem[{Sidi et~al.(2011)Sidi, Van~Kaick, Kleiman, Zhang, Cohen-Or
  et~al.}]{sidi2011unsupervised}
Sidi, O.; Van~Kaick, O.; Kleiman, Y.; Zhang, H.; Cohen-Or, D.; et~al. 2011.
\newblock Unsupervised co-segmentation of a set of shapes via descriptor-space
  spectral clustering.
\newblock In \emph{Proceedings of the 2011 SIGGRAPH Asia Conference}, 1--10.

\bibitem[{Su et~al.(2023)Su, Deng, Sun, Lin, Su, Wu et~al.}]{su2022unified}
Su, Y.; Deng, J.; Sun, R.; Lin, G.; Su, H.; Wu, Q.; et~al. 2023.
\newblock A Unified Transformer Framework for Group-based Segmentation:
  Co-Segmentation, Co-Saliency Detection and Video Salient Object Detection.
\newblock \emph{IEEE Transactions on Multimedia}, 1--13.

\bibitem[{Vaswani et~al.(2017)Vaswani, Shazeer, Parmar, Uszkoreit, Jones,
  Gomez, Kaiser, Polosukhin et~al.}]{vaswani2017attention}
Vaswani, A.; Shazeer, N.; Parmar, N.; Uszkoreit, J.; Jones, L.; Gomez, A.~N.;
  Kaiser, {\L}.; Polosukhin, I.; et~al. 2017.
\newblock Attention is all you need.
\newblock \emph{Advances in neural information processing systems}, 30.

\bibitem[{Vicente et~al.(2011)Vicente, Rother, Kolmogorov
  et~al.}]{vicente2011object}
Vicente, S.; Rother, C.; Kolmogorov, V.; et~al. 2011.
\newblock Object cosegmentation.
\newblock In \emph{CVPR 2011}, 2217--2224. IEEE.

\bibitem[{Wang et~al.(2017)Wang, Zhang, Yang, Cao, Xiong et~al.}]{Wang2017Mu}
Wang, C.; Zhang, H.; Yang, L.; Cao, X.; Xiong, H.; et~al. 2017.
\newblock Multiple Semantic Matching on Augmented $N$ -Partite Graph for Object
  Co-Segmentation.
\newblock \emph{IEEE Transactions on Image Processing}, PP: 1--1.

\bibitem[{Wang et~al.(2013)Wang, Huang, Guibas et~al.}]{wang2013image}
Wang, F.; Huang, Q.; Guibas, L.~J.; et~al. 2013.
\newblock Image co-segmentation via consistent functional maps.
\newblock In \emph{Proceedings of the IEEE international conference on computer
  vision}, 849--856.

\bibitem[{Xie et~al.(2023)Xie, Wang, Ma, Chen, Lu, Yang, Shi, Lin
  et~al.}]{xie2023edit}
Xie, D.; Wang, R.; Ma, J.; Chen, C.; Lu, H.; Yang, D.; Shi, F.; Lin, X.; et~al.
  2023.
\newblock Edit everything: A text-guided generative system for images editing.
\newblock \emph{arXiv preprint arXiv:2304.14006}.

\bibitem[{Xu et~al.(2021)Xu, Zhang, Wei, Lin, Cao, Hu, Bai
  et~al.}]{xu2021simple}
Xu, M.; Zhang, Z.; Wei, F.; Lin, Y.; Cao, Y.; Hu, H.; Bai, X.; et~al. 2021.
\newblock A simple baseline for zero-shot semantic segmentation with
  pre-trained vision-language model.
\newblock \emph{arXiv preprint arXiv:2112.14757}.

\bibitem[{Yuan et~al.(2017)Yuan, Lu, Wu et~al.}]{yuan2017deep}
Yuan, Z.-H.; Lu, T.; Wu, Y.; et~al. 2017.
\newblock Deep-dense Conditional Random Fields for Object Co-segmentation.
\newblock In \emph{IJCAI}, volume~1, 2.

\bibitem[{Zhang et~al.(2020{\natexlab{a}})Zhang, Cai, Lin, Shen
  et~al.}]{zhang2020deepemd}
Zhang, C.; Cai, Y.; Lin, G.; Shen, C.; et~al. 2020{\natexlab{a}}.
\newblock Deepemd: Few-shot image classification with differentiable earth
  mover's distance and structured classifiers.
\newblock In \emph{Proceedings of the IEEE/CVF conference on computer vision
  and pattern recognition}, 12203--12213.

\bibitem[{Zhang et~al.(2021)Zhang, Li, Lin, Wu, Yao
  et~al.}]{zhang2021cyclesegnet}
Zhang, C.; Li, G.; Lin, G.; Wu, Q.; Yao, R.; et~al. 2021.
\newblock Cyclesegnet: Object co-segmentation with cycle refinement and region
  correspondence.
\newblock \emph{IEEE Transactions on Image Processing}, 30: 5652--5664.

\bibitem[{Zhang et~al.(2020{\natexlab{b}})Zhang, Chen, Liu, Liu
  et~al.}]{zhang2020deep}
Zhang, K.; Chen, J.; Liu, B.; Liu, Q.; et~al. 2020{\natexlab{b}}.
\newblock Deep object co-segmentation via spatial-semantic network modulation.
\newblock In \emph{Proceedings of the AAAI Conference on Artificial
  Intelligence}, volume~34, 12813--12820.

\bibitem[{Zhang et~al.(2018)Zhang, Zhang, Peng, Xue, and Sun}]{zhang2018exfuse}
Zhang, Z.; Zhang, X.; Peng, C.; Xue, X.; and Sun, J. 2018.
\newblock Exfuse: Enhancing feature fusion for semantic segmentation.
\newblock In \emph{Proceedings of the European conference on computer vision
  (ECCV)}, 269--284.

\bibitem[{Zhou et~al.(2021)Zhou, Loy, Dai et~al.}]{zhou2021denseclip}
Zhou, C.; Loy, C.~C.; Dai, B.; et~al. 2021.
\newblock Denseclip: Extract free dense labels from clip.
\newblock \emph{arXiv preprint arXiv:2112.01071}.

\bibitem[{Zhou et~al.(2023)Zhou, Lei, Zhang, Liu, Liu et~al.}]{zhou2023zegclip}
Zhou, Z.; Lei, Y.; Zhang, B.; Liu, L.; Liu, Y.; et~al. 2023.
\newblock Zegclip: Towards adapting clip for zero-shot semantic segmentation.
\newblock In \emph{Proceedings of the IEEE/CVF Conference on Computer Vision
  and Pattern Recognition}, 11175--11185.

\bibitem[{Zou et~al.(2023)Zou, Yang, Zhang, Li, Li, Gao, Lee
  et~al.}]{zou2023segment}
Zou, X.; Yang, J.; Zhang, H.; Li, F.; Li, L.; Gao, J.; Lee, Y.~J.; et~al. 2023.
\newblock Segment everything everywhere all at once.
\newblock \emph{arXiv preprint arXiv:2304.06718}.

\end{thebibliography}

\end{document}